 \documentclass[pmlr,twocolumn,10pt]{jmlr} % W&CP article

\usepackage{booktabs}
\usepackage{siunitx}

\usepackage[switch]{lineno}

\usepackage{tcolorbox}
\usepackage{listings}
\usepackage{caption}
\usepackage{subcaption}
\usepackage{floatrow}
\definecolor{salmon}{rgb}{0.98, 0.5, 0.45}
\definecolor{royalblue}{rgb}{0.25, 0.41, 0.88}
\definecolor{darkblue}{rgb}{0.00, 0.00, 0.80}
\definecolor{boxbordercolor}{RGB}{239, 68, 68}
\definecolor{boxbackcolor}{RGB}{254, 226, 226}

\theorembodyfont{\upshape}
\theoremheaderfont{\scshape}
\theorempostheader{:}
\theoremsep{\newline}

 \title[Syn-STARTS: Synthesized START Triage Scenario Generation Framework]{Syn-STARTS: Synthesized START Triage Scenario Generation Framework for Scalable LLM Evaluation
}

\author{%
\Name{Chiharu Hagiwara}\textsuperscript{1,3} \Email{chiharu.hagiwara@riken.jp}\\
\Name{Naoki Nonaka}\textsuperscript{1,2} \Email{naoki.nonaka@riken.jp}\\
\Name{Yuhta Hashimoto}\textsuperscript{1} \Email{yuhta.hashimoto@riken.jp}\\
\Name{Ryo Uchimido}\textsuperscript{3} \Email{uchrccm@tmd.ac.jp}\\
\Name{Jun Seita}\textsuperscript{1,2} \Email{jun.seita@riken.jp}\\
\addr 1. Predictive Medicine Special Project, RIKEN Center for Integrative Medical Sciences, RIKEN, Tokyo, Japan \\
\addr 2. RIKEN Center for Interdisciplinary Theoretical and Mathematical Sciences, RIKEN, Tokyo, Japan \\
\addr 3. Faculty of Medicine, Institute of Science Tokyo, Tokyo, Japan
}

\begin{document}

\maketitle

\begin{abstract}
Triage is a critically important decision-making process in mass casualty incidents (MCIs) to maximize victim survival rates. While the role of AI in such situations is gaining attention for making optimal decisions within limited resources and time, its development and performance evaluation require benchmark datasets of sufficient quantity and quality. However, MCIs occur infrequently, and sufficient records are difficult to accumulate at the scene, making it challenging to collect large-scale real-world data for research use. Therefore, we developed Syn-STARTS, a framework that uses LLMs to generate triage cases, and verified its effectiveness. The results showed that the triage cases generated by Syn-STARTS were qualitatively indistinguishable from the TRIAGE open dataset generated by manual curation from training materials. Furthermore, when evaluating the LLM accuracy using hundreds of cases each from the green, yellow, red, and black categories defined by the standard triage method START, the results were found to be highly stable. This strongly indicates the possibility of synthetic data in developing high-performance AI models for severe and critical medical situations.
\end{abstract}


\paragraph*{Data and Code Availability}
The complete data and source code are available at GitHub \footnote{\url{https://github.com/seitalab/Syn-STARTS}}.

\paragraph*{Institutional Review Board (IRB)}
IRB approval was not required for this study as all evaluation data did not involve human subjects.

\section{Introduction}
Triage is a distinctive clinical task that demands the integration of multiple essential competencies central to high-stakes medical practice. Its primary aim is to maximize survival by prioritizing patients under the severe resource constraints of mass casualty incidents (MCIs) \citep{bazyar2019triage}. A widely adopted example is the START (Simple Triage and Rapid Treatment) algorithm, which assigns patients to one of four categories based solely on three parameters: respiration, circulation, and consciousness. Effective performance in this setting requires a specific combination of skills: strict adherence to algorithmic rules, rapid decision-making under extreme pressure, and equitable, bias-resistant judgment. These abilities are seldom assessed in combination in routine clinical practice, yet they epitomize the broader challenges any high-stakes medical AI must overcome. Accordingly, assessing a large language model’s (LLM) capacity for triage provides not only a rigorous benchmark of its technical proficiency but also a critical indicator of its readiness for meaningful and responsible clinical deployment.

Developing a reliable method to evaluate the triage capabilities of LLMs requires benchmark datasets with appropriate quality and quantity, yet using data from real-world MCIs presents formidable challenges. Empirical MCI events are both rare and difficult to be documented at the site. It is also known that available empirical datasets are often compromised by incompleteness, ambiguity, or bias \citep{ghassemi2020review}. Furthermore, utilizing real patient data incurs substantial acquisition costs and poses privacy and ethical concerns. To overcome such limitations and to establish a more scalable and comprehensive assessment of the triage capabilities of LLMs, we introduce Syn-STARTS, a framework that systematically generates and validates synthetic triage cases using LLMs.

\section{Related Work}
START is the most widely adopted global standard method for triage \citep{start_proposal,franc2022metastart,iserson2007triage,benson1996disaster}. It is explicitly designed to maximize the number of survivors under the limited resource conditions of an MCI by prioritizing treatment for the most urgent, salvageable patients. START assign patients to one of four categories: Immediate (red), Delayed (yellow), Minor (green), and Deceased/Expectant (black), based on sequential evaluation of respiratory status, circulatory adequacy, and consciousness. By confining assessment to these objective criteria, START minimizes subjective judgment and sociocultural bias, thereby supporting equitable and utilitarian patient prioritization \citep{iserson2007triage}. Its global prominence and continued clinical use make it an appropriate foundation for this study.

Recent work has begun to evaluate LLMs within the START framework \citep{kirch2024triage, franc2024accuracy}. \citet{franc2024accuracy} made a total of 380 simulated patients from the Canadian disaster database, and examined ChatGPT’s triage performance through repeated tag predictions under varied prompts, assessing both accuracy and stability. Unfortunately, the simulated patient cases are not publicly shared. \citet{kirch2024triage} manually curated 87 patient descriptions from START and Jump-START\footnote{The Jump-START protocol is an adaptation of the START triage system for the physiological characteristics of pediatric patients.} training materials and introduced a public dataset called \textit{TRIAGE,}\footnote{Available at \url{https://huggingface.co/datasets/NLie2/TRIAGE}.} and evaluated six LLMs. The TRIAGE dataset contains 54 adult cases and 33 pediatric cases, and tag distributions are \{Green: 18, Yellow: 11, Red: 22, Black: 3\}, and \{Green: 7, Yellow: 11, Red: 11, Black: 4\}, respectively. Hereafter, triage tag counts are presented in the fixed order of \{Green, Yellow, Red, Black\}.

Leveraging LLMs to benchmark other LLMs is an innovative strategy to circumvent the limitations of evaluation datasets \citep{zhu2024dynamic,loni2025review,auto2025kim}. In the context of medicine, this approach directly addresses the critical issues of real-patient data scarcity, high acquisition costs, and patient privacy \citep{loni2025review}. Although this strategy is promising, it also introduces unique challenges, including data contamination from model-specific biases \citep{xu2024pride} and the risk of diminished data diversity from ``model collapse'' \citep{shumailov2024curse}. Acknowledging these risks, prior work has explored strategies to verify synthetic datasets, including validation by human experts \citep{cheung2023chatgpt} and algorithmic validity checks \citep{patel2025llmgen,kashyap2025struct}.

\section{The Syn-STARTS Framework}
We propose Syn-STARTS, a framework that systematically generates and validates synthetic triage cases. Its atomic output is a structured data object referred to as a ``Syn-STARTS case.''

\subsection{Structure of Syn-STARTS Case}\label{sec:gen-and-val}
The Syn-STARTS case utilizes a specific data schema, which comprises three core elements: a ground-truth triage \texttt{tag}, a structured set of \texttt{vitals}, and a narrative \texttt{description} of the patient’s condition (exemplified in \figureref{fig:1}). The \texttt{vitals} are designed to serve as a verifiable, intermediate layer that both anchors the narrative and enables deterministic validation of the triage tag.

\begin{figure}[htbp]
\floatconts
  {fig:1}
  {\caption{The structure of a Syn-STARTS case, comprising the \texttt{tag}, structured \texttt{vitals}, and narrative \texttt{description}. The \texttt{vitals} provide a verifiable intermediate layer that links the \texttt{description} to the triage \texttt{tag} via algorithmic confirmation.}}
  {\includegraphics[width=0.95\linewidth]{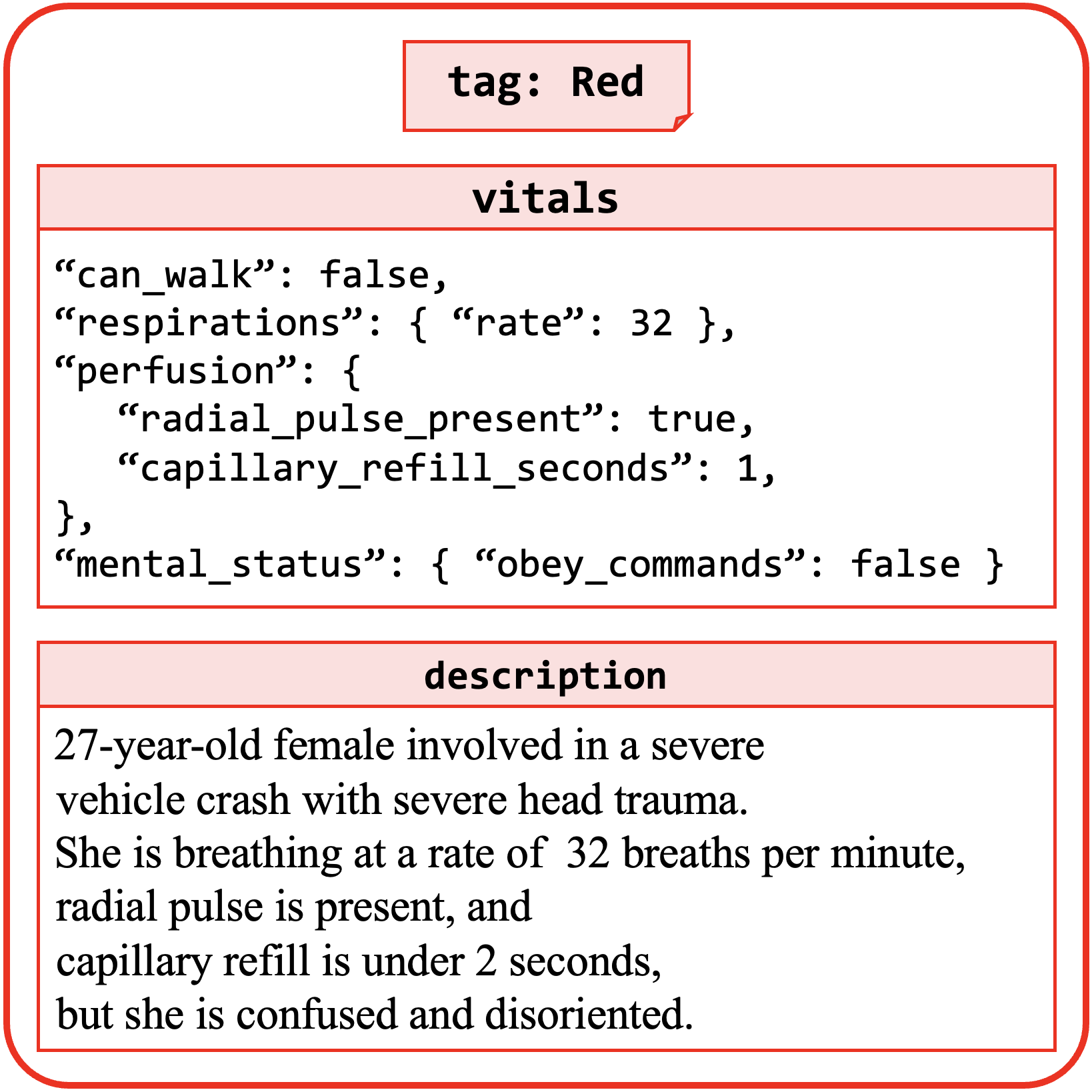}}
\end{figure}

\subsection{Syn-STARTS Case Synthesis Pipeline} \label{sec:case}
The Syn-STARTS framework synthesizes each Syn-STARTS case through its \textbf{generation} and \textbf{validation.}

\begin{enumerate}
    \item \textbf{Generation} \label{sec:generation}: A candidate case is generated using Llama-3.1-70B-Instruct, a leading open-source model selected to ensure independence from the models under evaluation. The prompt instructs the model to create a JSON object adhering to the tripartite data schema (\texttt{tag, vitals, description}) and the START algorithm's decision-making principles. This process is designed to yield a consistent and clinically plausible initial output. The prompts are detailed in \appendixref{apd:prompt} (\figureref{fig:8,fig:9}).

    \item \textbf{Validation}: Each generated candidate case is then subjected to a rigorous, fully automated, three-step validation pipeline: (1) \textbf{START Consistency}, which programmatically verifies the tag against the vitals; (2) \textbf{Medical Plausibility}, which checks vitals against predefined clinical constraints (e.g., ambulatory patients must have a palpable radial pulse); and (3) \textbf{Narrative Consistency}, which ensures the description is concordant with the vitals.
\end{enumerate}

Only cases that successfully pass all three validation stages are retained and certified as a ``Syn-STARTS case.'' Examples are presented in \appendixref{apd:example-cases}.

\section{Experiments}
To evaluate Syn-STARTS through a triage tag prediction task performed by a suite of LLMs, we present two main analyses: experiments validating that our synthetic datasets serve as a high-fidelity proxy for the ``TRIAGE'' existing expert-authored benchmark, and in-depth analyses that leverage our framework's unique capabilities.

\subsection{Experimental Setup} \label{sec:setup}
\paragraph{Corpus Construction.} \label{sec:corpus}
To serve as the basis for our experiments, we first utilized the Syn-STARTS framework to construct a large, foundational corpus of 2,000 validated cases, uniformly distributed across the four triage tags, comprising 500 cases for each tag. The processes are summarized in \appendixref{apd:algo}. This corpus serves as the central data pool from which the synthetic experimental datasets were derived.

\paragraph{Datasets.} \label{sec:chara}
We prepared two dataset types (visualized in \figureref{fig:2}): (1) the existing expert-authored ``TRIAGE-adult dataset'' derived from the TRIAGE benchmark \citep{kirch2024triage} and (2) our synthetic ``Syn-STARTS datasets'': composed of the ground-truth \texttt{tags} and narrative \texttt{descriptions} of Syn-STARTS cases sampled from the 2,000-case corpus. To enable variance analysis in our results, each Syn-STARTS configuration described below consists of ten non-overlapping replicate datasets, created by sampling from the corpus without replacement (see \appendixref{apd:chara} for a linguistic analysis).

\begin{figure}[htbp]
\floatconts
  {fig:2}
  {\caption{The experimental datasets include the TRIAGE-adult dataset and Syn-STARTS datasets, variants configured by tag distribution and scale, with each configuration comprising ten non-overlapping replicate datasets without replacement. Hereafter, triage tag counts are presented in the fixed order of \{Green, Yellow, Red, Black\}.}}
  {\includegraphics[width=\linewidth]{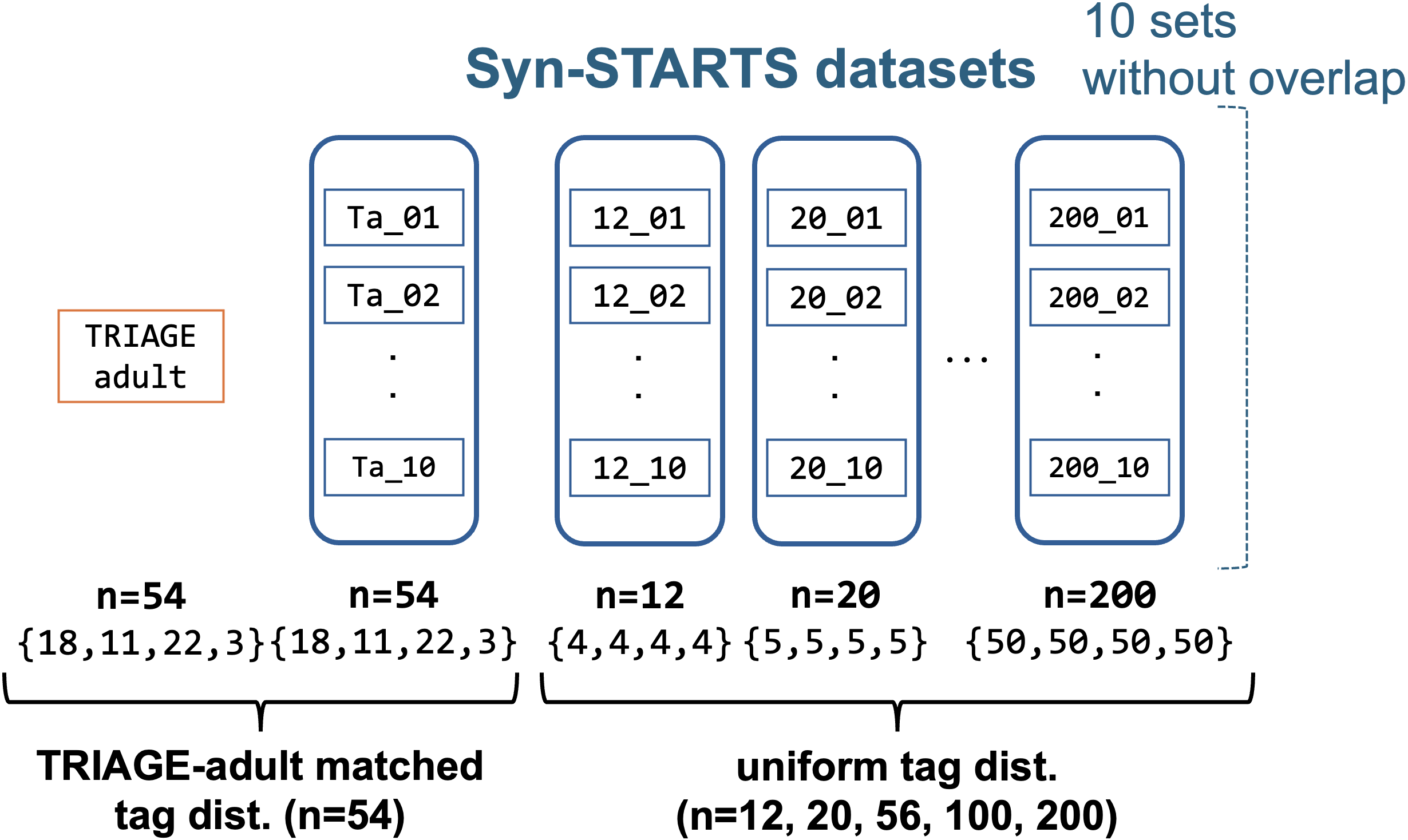}}
\end{figure}

\begin{enumerate}
    \item \textbf{TRIAGE-adult Dataset} ($n=54$; $\{18, 11, 22, 3\}$): It comprises the START-compliant (adult) cases from the original TRIAGE benchmark; we excluded pediatric (Jump-START) cases to focus the analysis on a single protocol.
    \item \textbf{Syn-STARTS Datasets}: We sampled cases from the corpus along two experimental axes:
    \begin{enumerate}
        \item TRIAGE-adult matched tag distribution ($n=54$; $\{18, 11, 22, 3\}$).
        \item Uniform tag distribution ($\{n/4, n/4, n/4, n/4\}$), of varying scales ($n \in \{12, 20, 56, 100, 200\}$).
    \end{enumerate}
\end{enumerate}

\paragraph{Models.} Following \citet{kirch2024triage}, we evaluated a suite of six models (two open-source and four closed-source): Mistral-7B-Instruct (hereafter, Mistral 7B) \citep{jiang2023mistral}, Mixtral-8x22B-Instruct (hereafter, Mixtral 8x22B) \footnote{\url{https://mistral.ai/news/mixtral-8x22b}}, Claude 3 Haiku\footnote{\url{https://www.anthropic.com/news/claude-3-haiku}}, Claude Opus 4\footnote{\url{https://www.anthropic.com/news/claude-4}}: as the Claude 3 Opus version used in prior work was unavailable, we used its successor model per Anthropic's recommendation, GPT-3.5 \footnote{\url{https://platform.openai.com/docs/models/gpt-3.5-turbo}}, and GPT-4 \footnote{\url{https://platform.openai.com/docs/models/gpt-4-turbo}}.

\paragraph{Task and Evaluation.} \label{sec:task-eval} Our methodology adhered to prior work \citep{kirch2024triage}. Models were provided with a patient's narrative description and prompted to output the corresponding START triage tag (see \appendixref{apd:prompt}; \figureref{fig:10} for prompt details). Accuracy, defined as the proportion of correct predictions against the ground truth, served as the primary evaluation metric.

\subsection{Comparison with TRIAGE-adult}
To validate the Syn-STARTS datasets, we assessed them from two perspectives: \textbf{perceptual realism} in comparison to expert judgment and \textbf{performance fidelity} in relation to the established expert-authored benchmark.

First, to rigorously evaluate the perceptual realism of the Syn-STARTS cases, we designed and conducted a blinded comparative judgment experiment involving three medical doctors. In this blinded A/B test, they independently performed a forced-choice task over a total of 20 questions, identifying which of the two presented cases they believed was synthetically generated. Each question presented a randomized pair of cases: one Syn-STARTS case from the corpus and one case from the TRIAGE-adult dataset. To ensure a fair comparison based on narrative plausibility rather than the clinical perspective, each pair was matched by the same ground-truth triage tag, and the presentation order within pairs was randomized. This overall design aimed to quantitatively measure the indistinguishability of our synthetic data from human-authored content.

Second, to assess the performance fidelity to the established benchmark, we measured the concordance of model performance between the TRIAGE-adult dataset and the Syn-STARTS datasets. Concordance was quantified using the Pearson correlation coefficient, calculated by correlating each model’s accuracy on the TRIAGE-adult dataset with its mean accuracy across the ten Syn-STARTS datasets possessing matched tag distributions ($n=54$; $\{18, 11, 22, 3\}$).

\subsection{Dataset Composition and Scale Analyses} \label{sec:method2}
Leveraging the unique scalability and controllability of the Syn-STARTS framework, we investigated the sensitivity of model evaluation to key dataset characteristics through three specific analyses: the \textbf{effect of dataset tag distribution,} the \textbf{effect of dataset scale,} and \textbf{model-specific tag sensitivity analysis.}

First, to investigate the effect of dataset tag distribution on evaluation outcomes, we compared model accuracies between ten Syn-STARTS datasets with non-uniform ``TRIAGE-adult'' tag distribution ($n=54$; $\{18, 11, 22, 3\}$) and ten Syn-STARTS datasets with a uniform tag distribution ($n=56$; $\{14, 14, 14, 14\}$), using a Wilcoxon signed-rank test for statistical significance.

Next, to assess the impact of dataset scale on evaluation stability, we measured the standard deviation of model accuracy across the ten replicate datasets for each scale tier ($n \in \{12, 20, 56, 100, 200\}$).

Finally, to identify model-specific tag sensitivity patterns, we constructed an averaged confusion matrix for each model from the largest datasets ($n=200$).

\section{Results} \label{sec:results}
The numerical data underlying the figures in this section are presented in \appendixref{apd:tables}.
% \paragraph{Overview of the Experiments.} Here we present the results of our two-part experiments. First, we confirmed its validity by its perceptual realism and established that our Syn-STARTS datasets serve as a high-fidelity proxy for the expert-authored benchmark. Second, we showcased the advanced in-depth analyses enabled by our framework's controllability and scalability.

\subsection{Comparison with TRIAGE-adult}

\paragraph{Perceptual Realism.} Our first question was whether our synthetic cases are perceptually indistinguishable from those authored by experts. To test this, we conducted a blinded test where three clinical experts were asked to distinguish our Syn-STARTS case from the TRIAGE-adult case. We found that the experts could not distinguish the synthetic cases, confirming the high perceptual realism of our generated data (\figureref{fig:3}).

\begin{figure}[htbp]
\floatconts
  {fig:3}
  {\caption{Expert discrimination of synthetic Syn-STARTS cases from expert-authored vignettes. \figureref{fig:3a}: Number of questions answered correctly by each expert; \figureref{fig:3b}: Averaged confusion matrix from the three experts, indicating Syn-STARTS cases are difficult to distinguish from expert-authored cases.}}
  {%
    \subfigure[]{\label{fig:3a}%
      \includegraphics[width=0.55\linewidth]{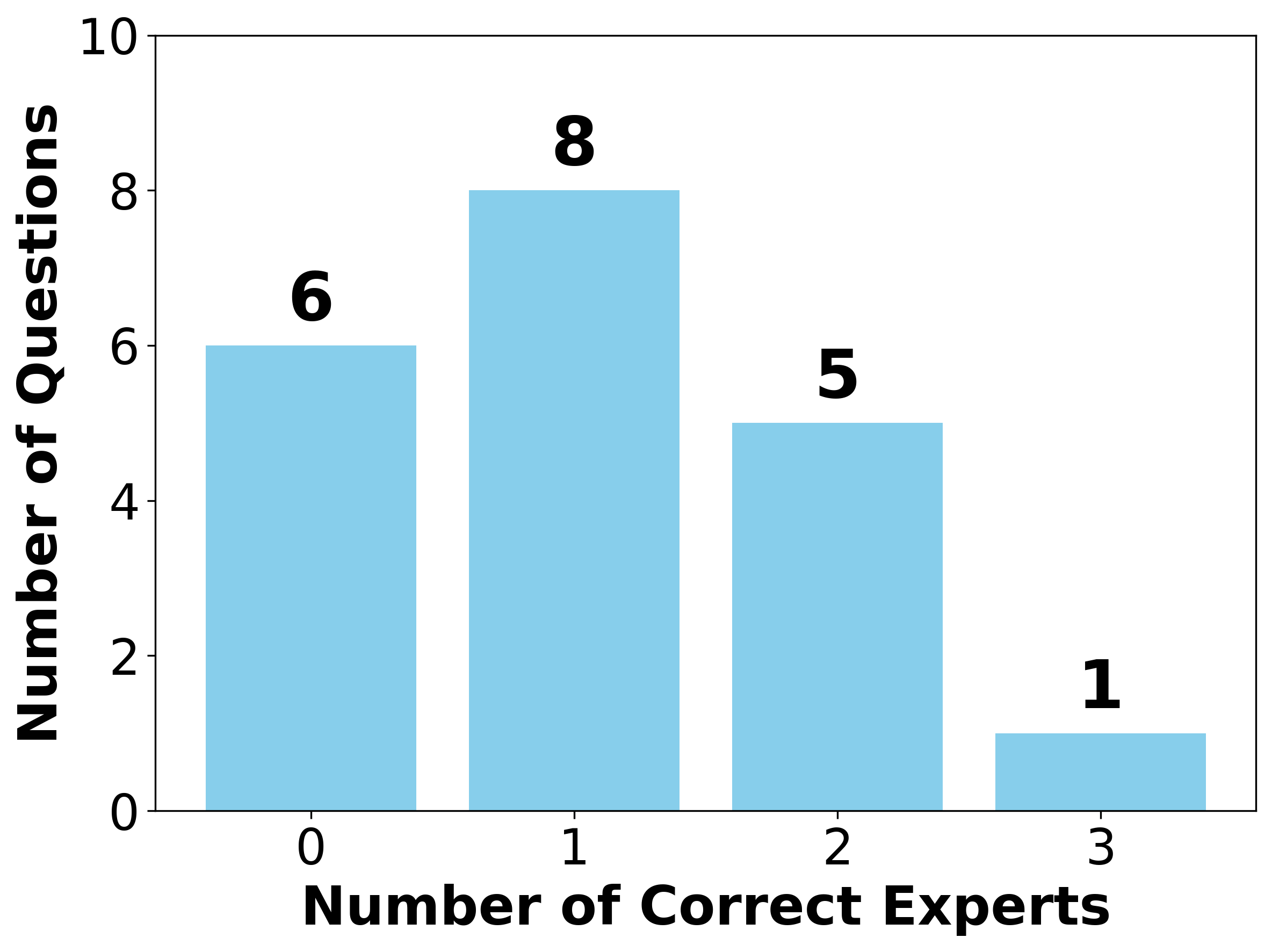}}%
    % \qquad
    \subfigure[]{\label{fig:3b}%
      \includegraphics[width=0.45\linewidth]{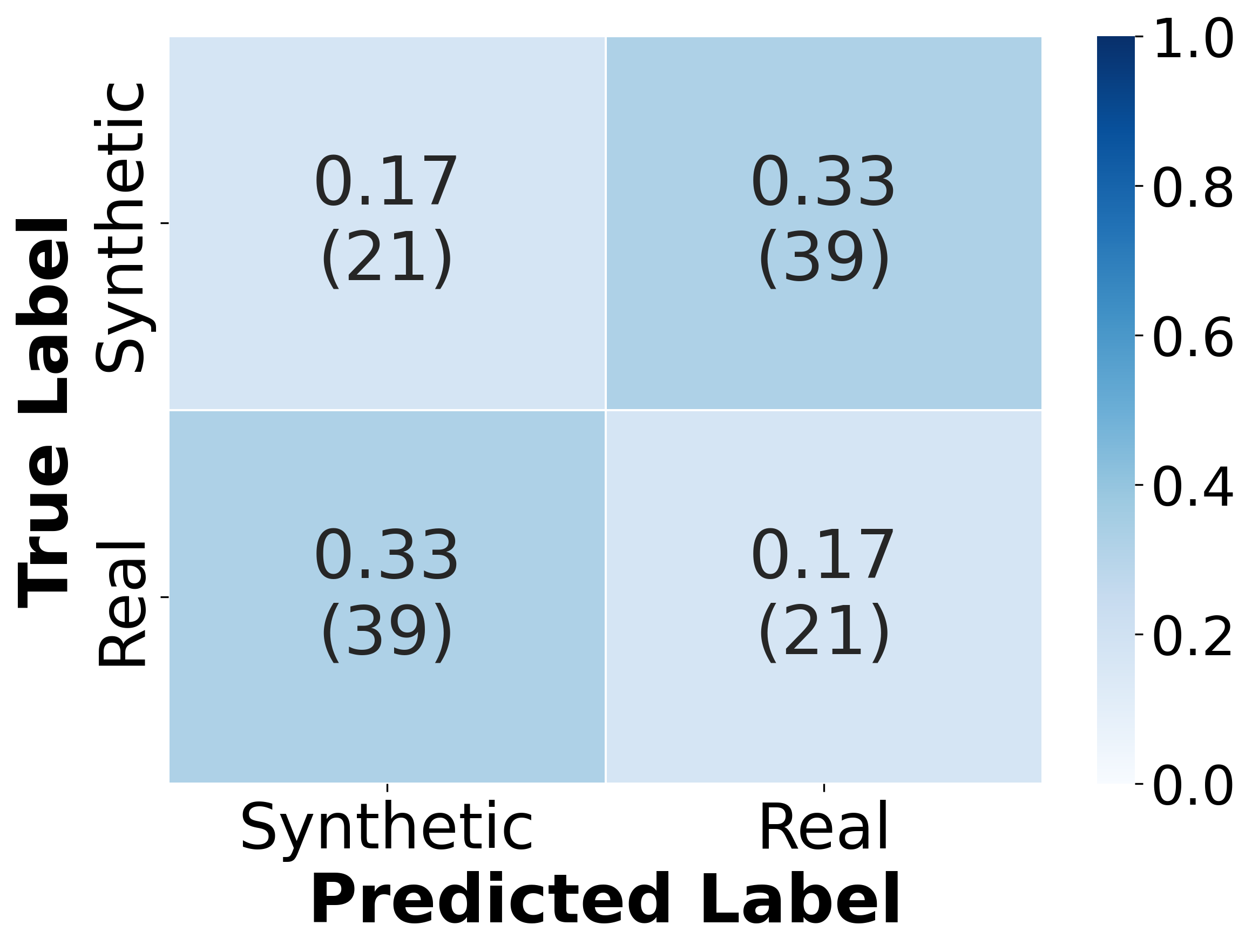}}
  }
\end{figure}

\paragraph{Performance Fidelity.} \label{sec:fidelity} Next, we aimed to assess the performance fidelity to the established expert-authored benchmark. We measured this by calculating the Pearson correlation between model accuracies on the TRIAGE-adult dataset and our Syn-STARTS datasets with a matched tag distribution ($n = 54$; $\{18, 11, 22, 3\}$). We found a strong positive correlation (Pearson’s $r = 0.92$, $p < 0.01$), confirming that Syn-STARTS datasets serve as a high-fidelity proxy for the expert-authored benchmark (\figureref{fig:4}).

\begin{figure}[htbp]
\floatconts
  {fig:4}
  {\caption{Scatter plot showing the relationship in model accuracy across the TRIAGE-adult and Syn-STARTS datasets, with matched tag distribution ($n = 54$; $\{18, 11, 22, 3\}$). The strong linear relationship (Pearson’s $r = 0.92$) suggests that performance is largely preserved across them.}}
  {\includegraphics[width=0.75\linewidth]{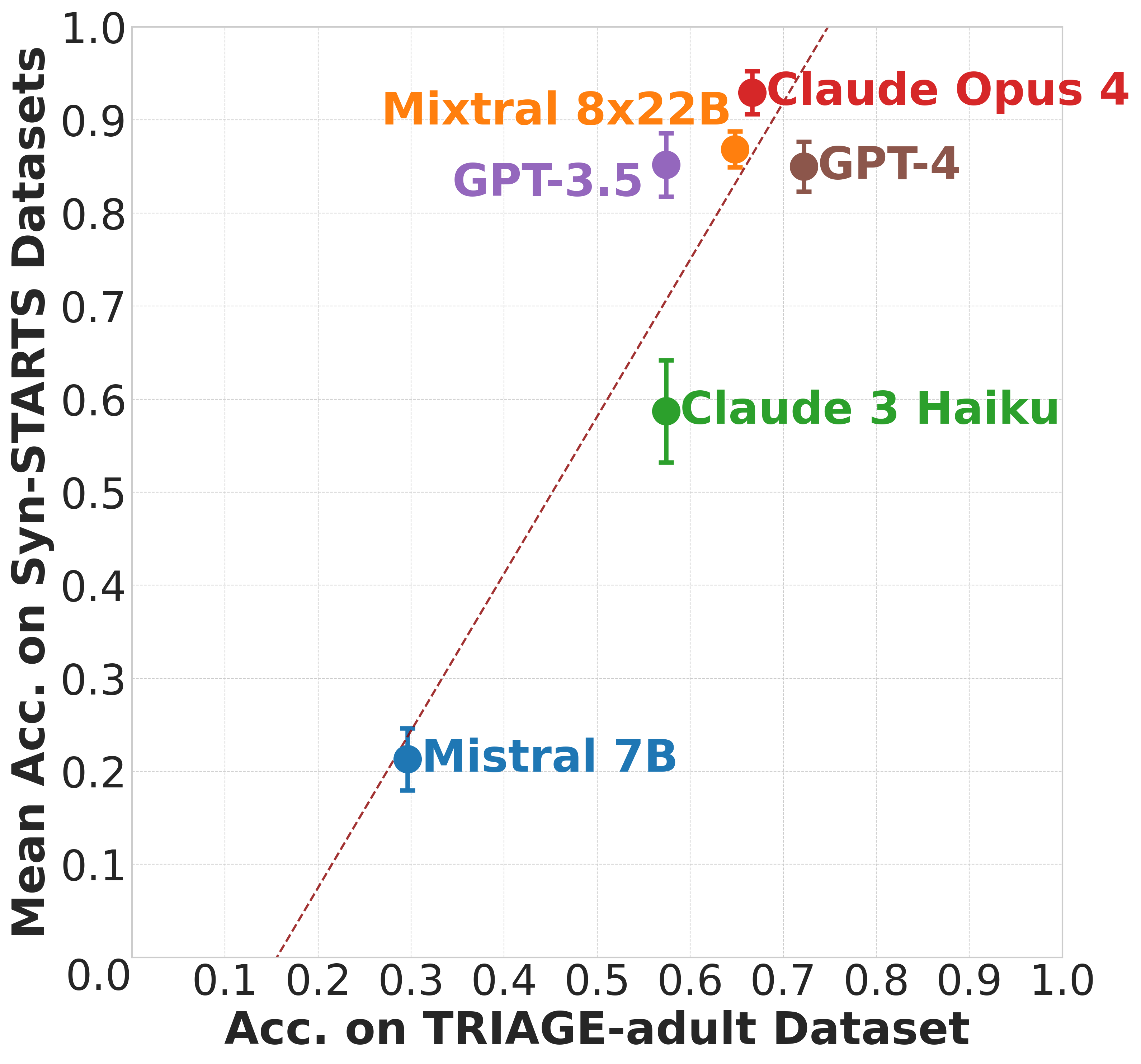}}
\end{figure}

\subsection{Dataset Composition and Scale Analyses}
Having validated the datasets, we leveraged the framework to conduct in-depth analyses, allowing us to: (1) reveal how dataset characteristics like tag distribution affect evaluation outcomes, (2) achieve more stable evaluations, and (3) unlock model-specific tag sensitivity analysis.

\paragraph{Effect of Tag Distribution.} \label{sec:tagdist} First, we investigated whether the evaluation dataset’s composition, specifically its distribution of triage tags, influences model performance outcomes. We compared model accuracies between datasets of similar scale; one with a ``TRIAGE adult'' tag distribution ($n = 54$; $\{18, 11, 22, 3\}$) and the other with a uniform tag distribution ($n = 56$; $\{14, 14, 14, 14\}$). A Wilcoxon signed-rank test revealed model-dependent sensitivity: GPT-3.5 ($p < 0.001$) and GPT-4 ($p = 0.04$) exhibited statistically significant declines in accuracy under uniform tag distributions. It indicates that dataset composition can be a critical, model-dependent factor that can skew evaluation results (\figureref{fig:5}).

\begin{figure*}[hbtp]
\floatconts
  {fig:5}
  {\caption{Accuracy distributions across Syn-STARTS datasets with ``TRIAGE adult'' tag distributions ($n=54$; $\{18, 11, 22, 3\}$) and uniform tag distributions ($n=56$; $\{14, 14, 14, 14\}$). The results reveal a model-dependent sensitivity to the dataset's composition. * denotes a statistically significant difference between the two distributions (Wilcoxon signed-rank test; GPT-3.5, $p < 0.01$; 
  GPT-4, $p = 0.04$).}}
  {\includegraphics[width=0.9\linewidth]{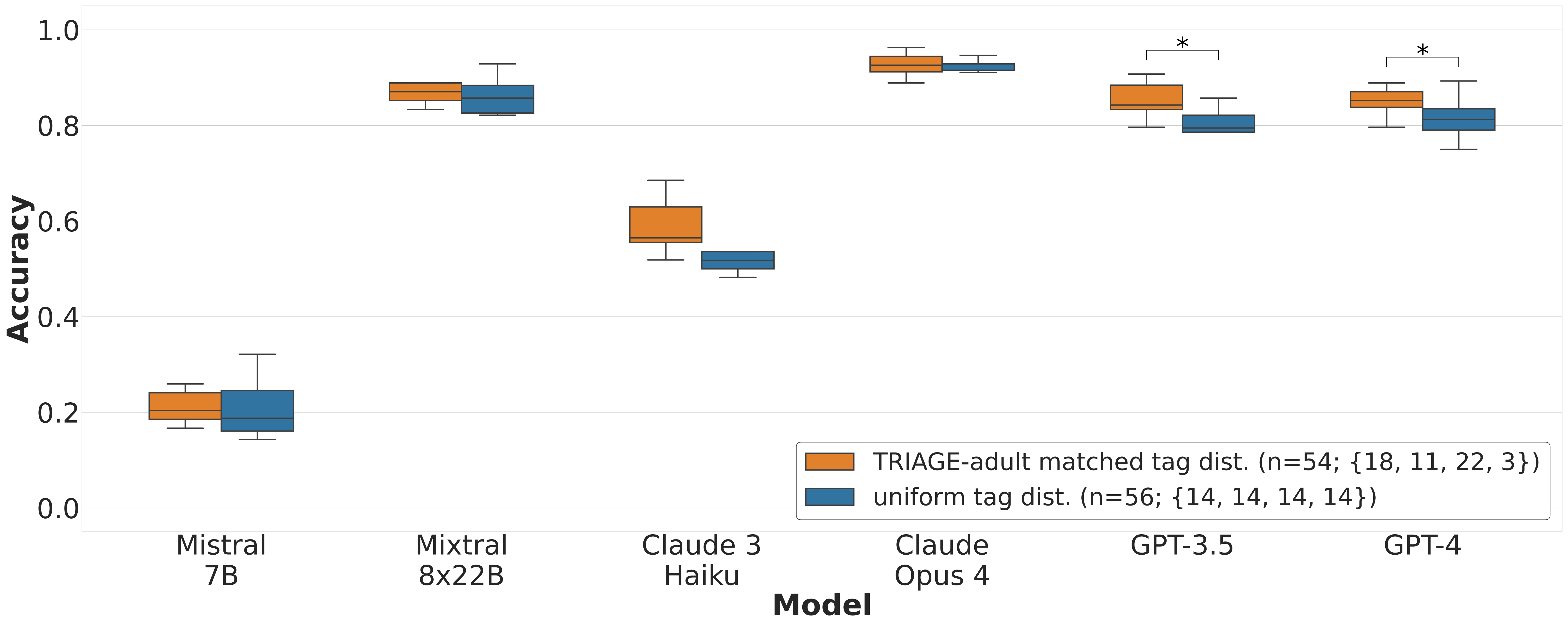}}
\end{figure*}

\paragraph{Effect of Dataset Scale.} \label{sec:scale}
Subsequently, we assessed the influence of dataset scale on the stability of the evaluation. This analysis was conducted by measuring the standard deviation of model accuracy across datasets of increasing scale ($n = 12, 20, 56, 100, \text{and } 200$) with a uniform distribution of tags. As shown in \figureref{fig:6}, the standard deviation of accuracy consistently diminished as the dataset size increased. This indicates that larger datasets yield more reliable evaluation results. Similar patterns were observed by tag (\appendixref{apd:tags}).

\begin{figure*}[htbp]
\floatconts
  {fig:6}
  {\caption{Standard deviation of model across datasets of increasing scale ($n = 12, 20, 56, 100, \text{and } 200$) with uniform tag distribution. The consistent downward trend shows that increasing dataset scale reduces variance in accuracy estimates.}}
  {\includegraphics[width=\linewidth]{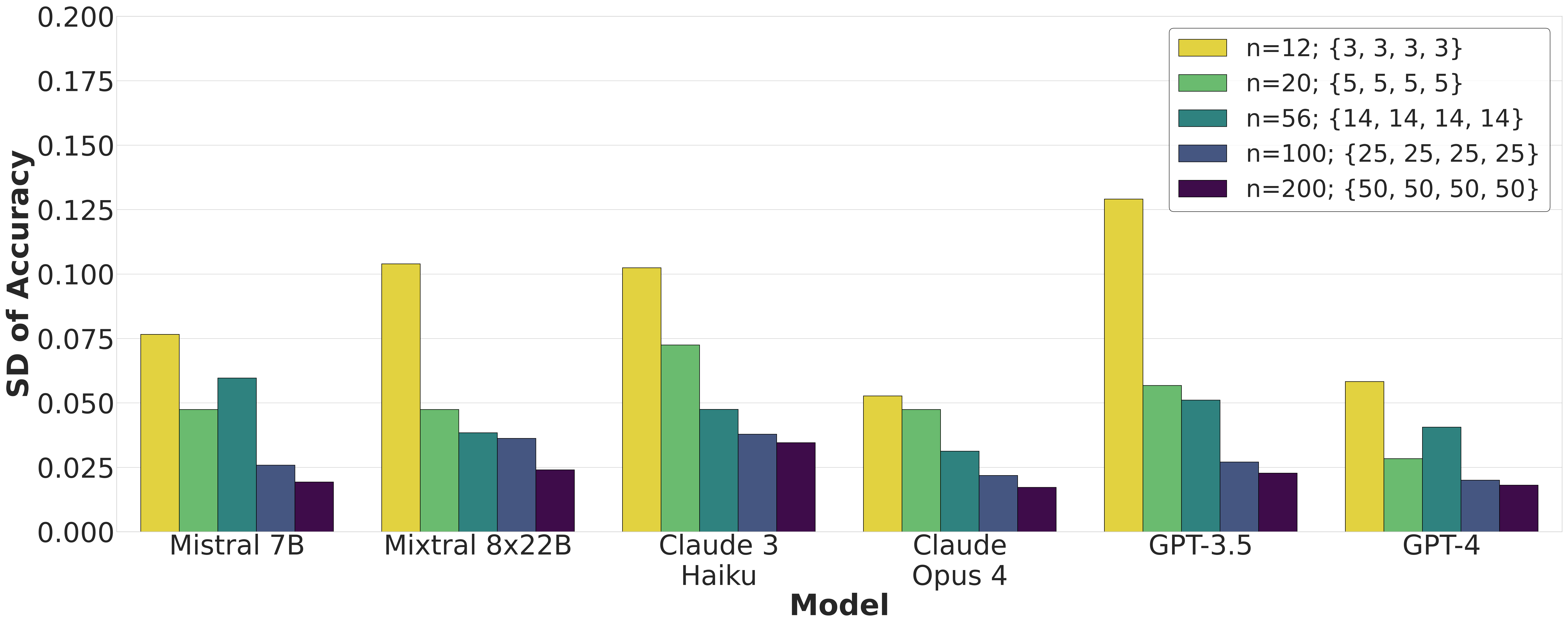}}
\end{figure*}

\paragraph{Model-Specific Tag Sensitivity Analysis.} Finally, to identify model-specific error patterns, we construct averaged confusion matrices for each model using our largest, most stable datasets ($n=200$). This analysis revealed that each model exhibits a distinct misclassification profile, uncovering unique strengths and weaknesses that would be obscured in smaller-scale evaluations (\figureref{fig:7}).

\begin{figure*}[htbp]
\floatconts
  {fig:7}
  {\caption{Averaged confusion matrices for each model based on Syn-STARTS datasets ($n=200$; $\{50, 50, 50, 50\}$). The differing misclassification patterns reveal model-specific error tendencies across triage categories.}}
  {
    \begin{minipage}[c]{0.9\linewidth}
      \centering
        \subfigure[Mistral 7B]{\label{fig:7a}
          \includegraphics[width=0.29\linewidth]{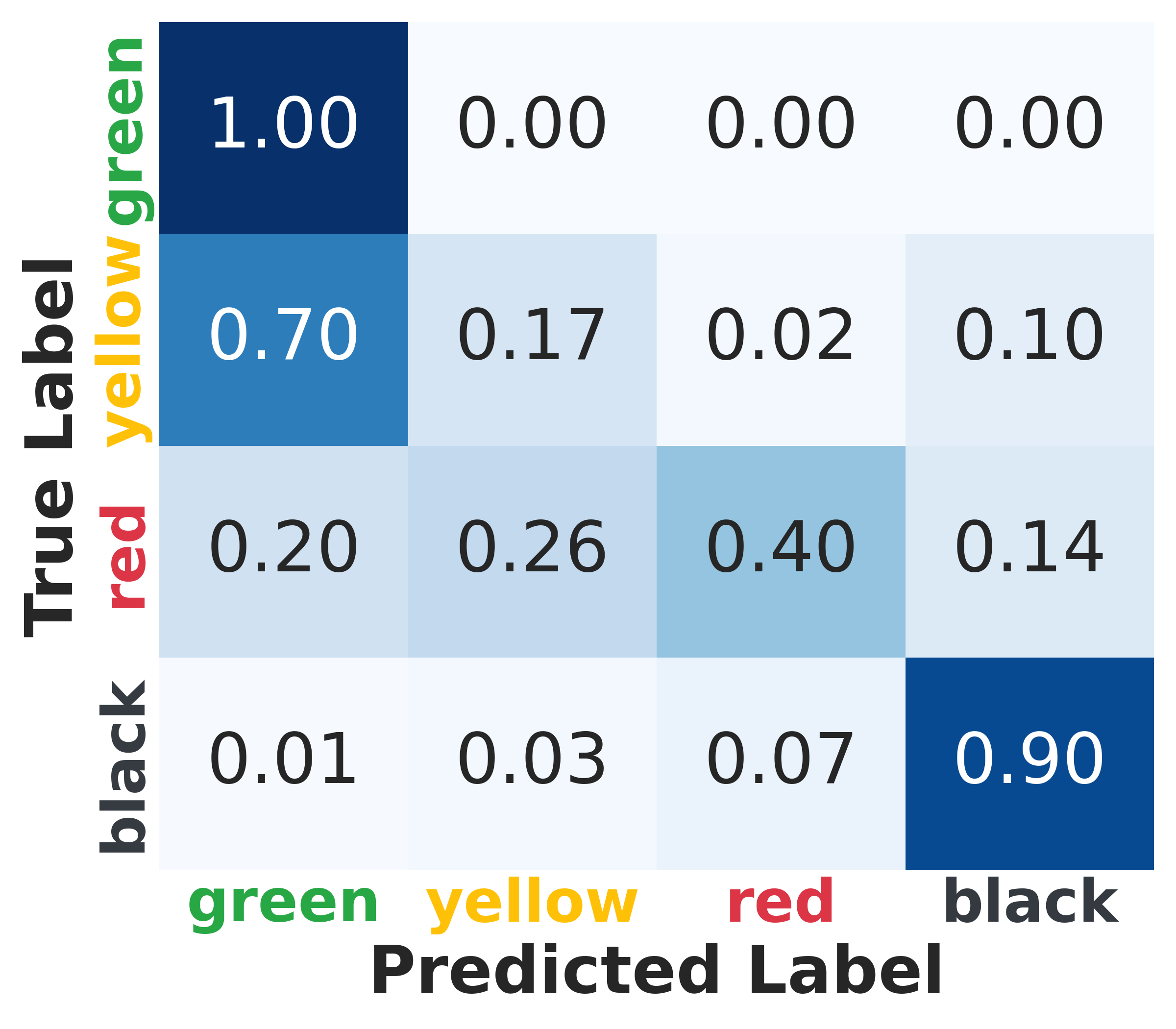}}
        \hfill
        \subfigure[Mixtral 8x22B]{\label{fig:7b}
          \includegraphics[width=0.29\linewidth]{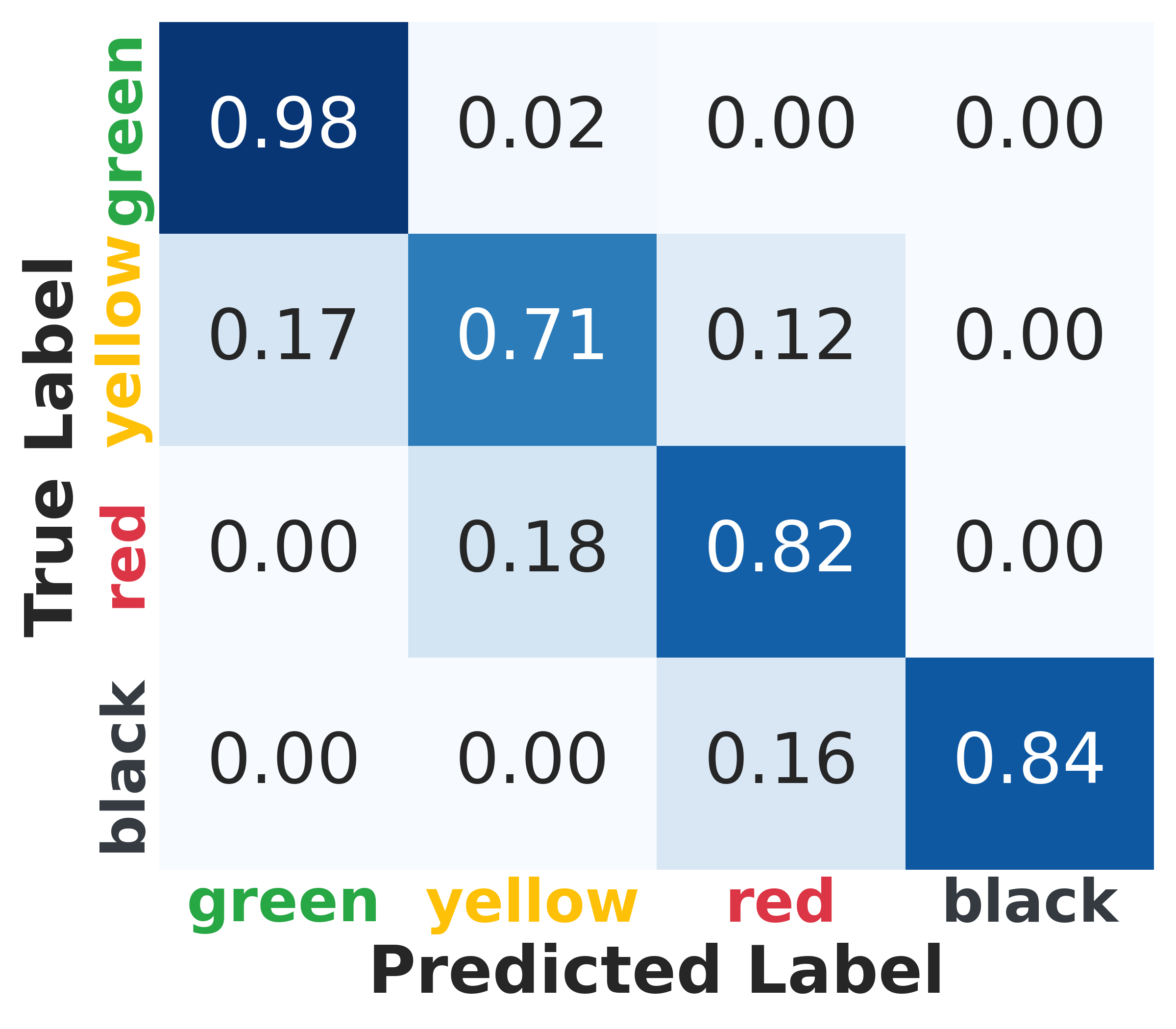}}
        \hfill
        \subfigure[Claude 3 Haiku]{\label{fig:7c}
          \includegraphics[width=0.29\linewidth]{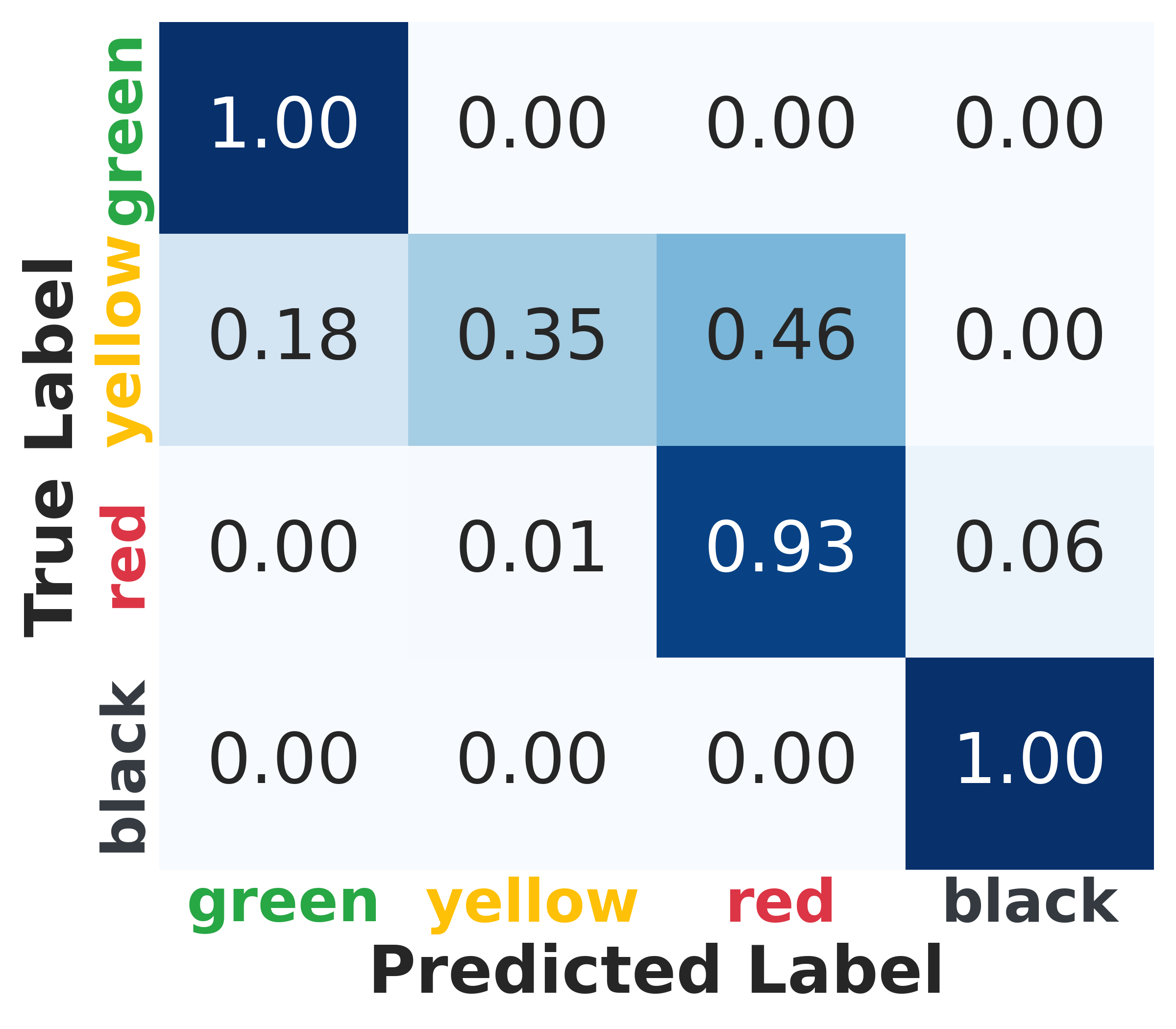}}
        \\
        \subfigure[Claude Opus 4]{\label{fig:7d}
          \includegraphics[width=0.29\linewidth]{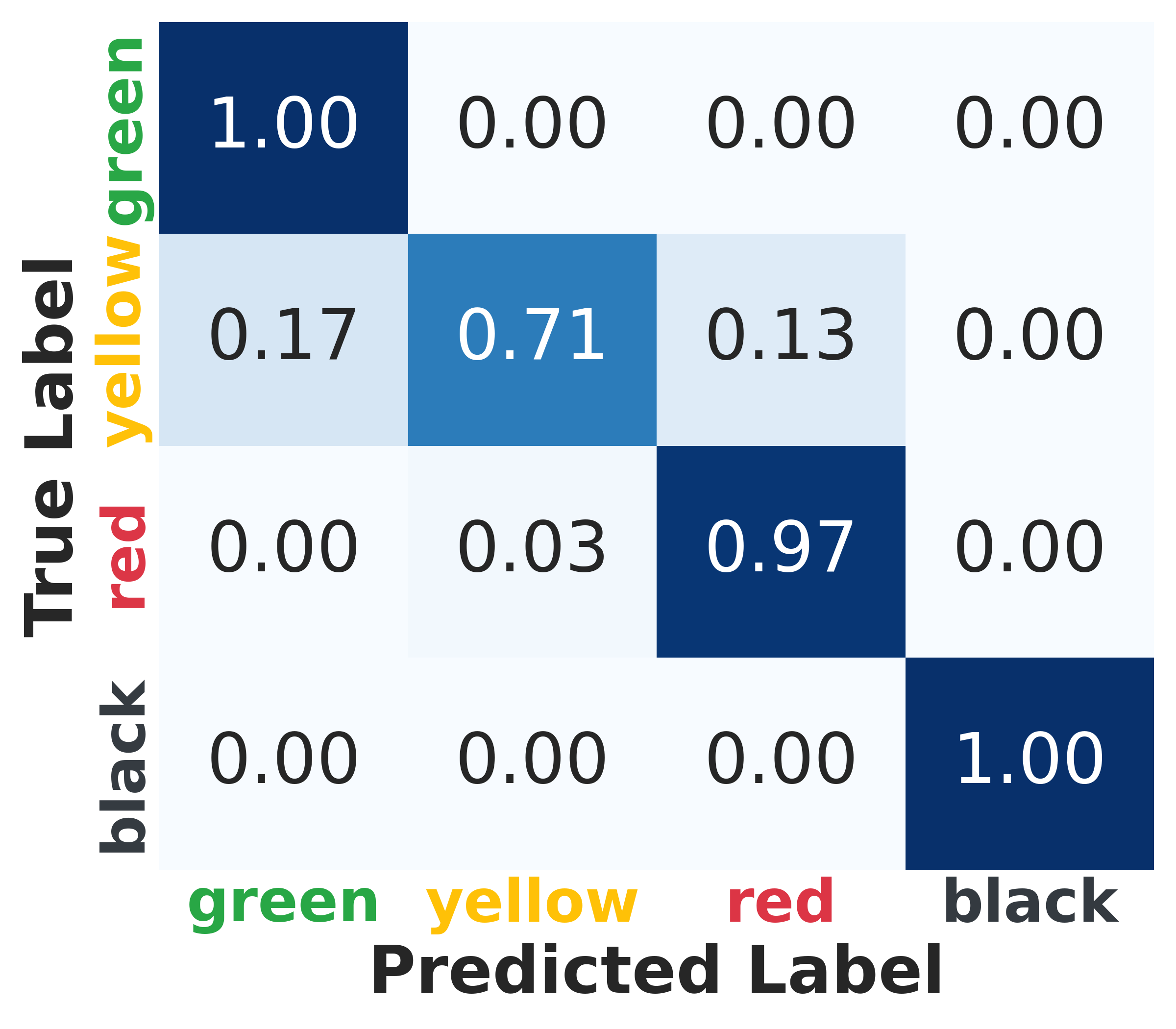}}
        \hfill
        \subfigure[GPT-3.5]{\label{fig:7e}
          \includegraphics[width=0.29\linewidth]{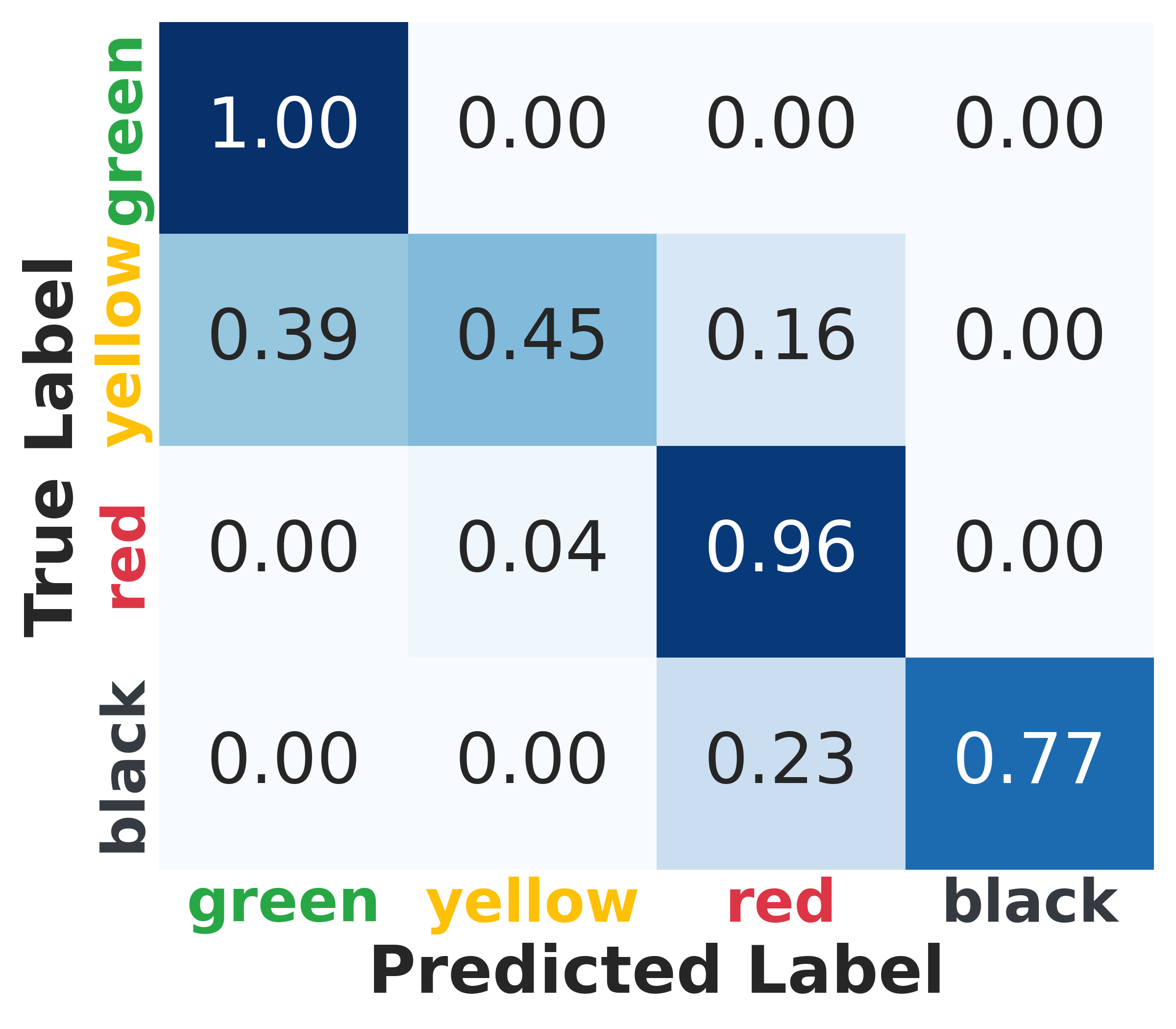}}
        \hfill
        \subfigure[GPT-4]{\label{fig:7f}
          \includegraphics[width=0.29\linewidth]{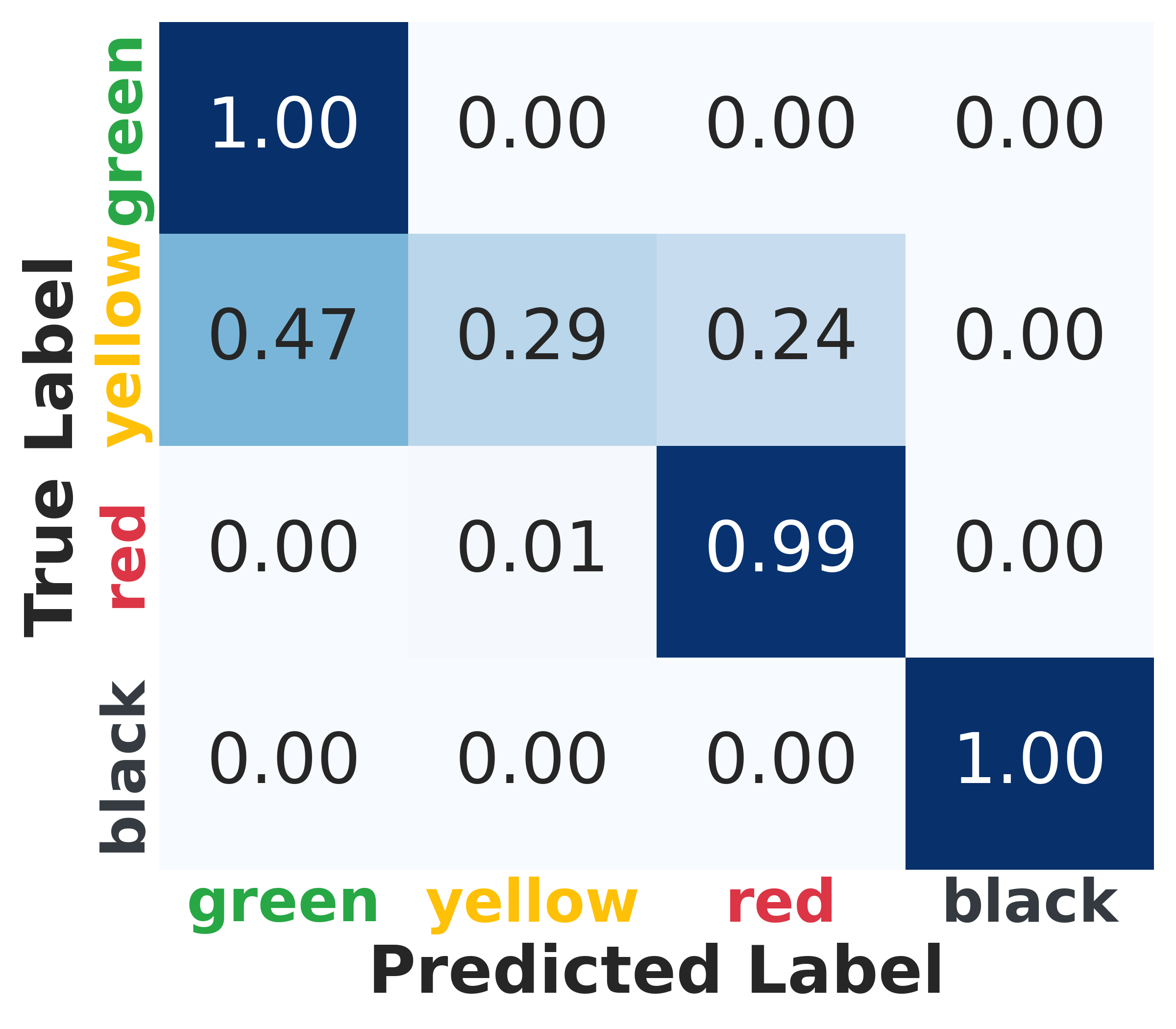}}
    \end{minipage}%
    \hfill
    \begin{minipage}[c]{0.08\linewidth}
      \includegraphics[width=0.8\linewidth]{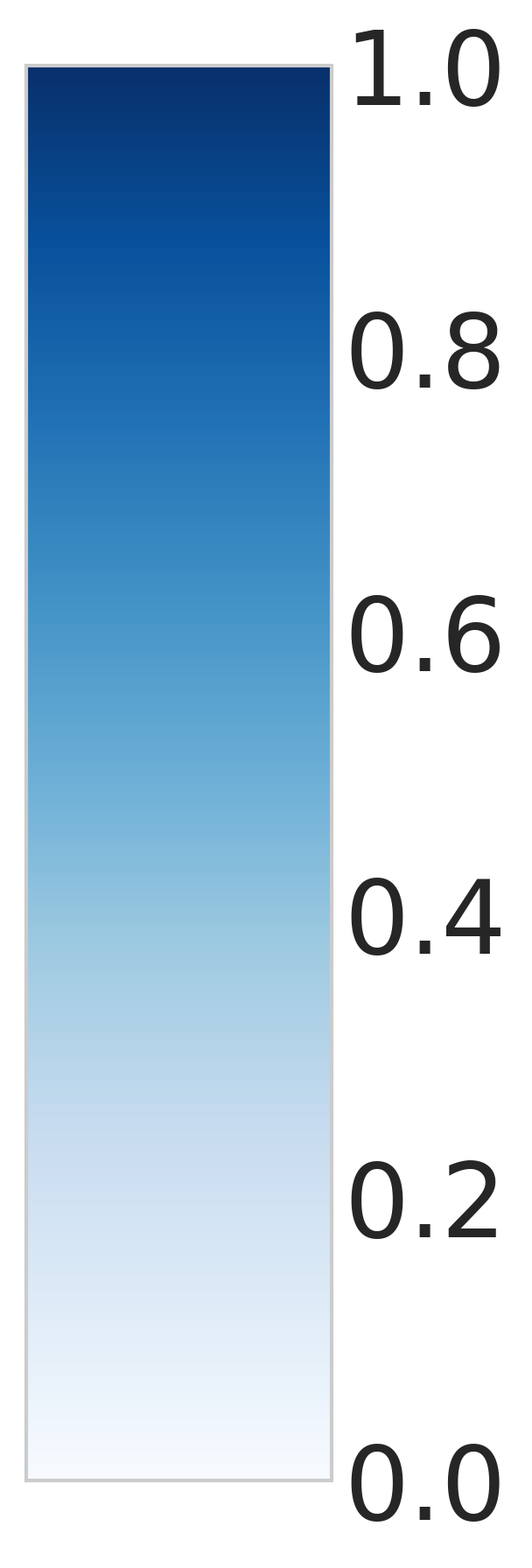}
    \end{minipage}
  }
\end{figure*}

\section{Discussion}
\subsection{Principal Findings and Implications}
Our results first show that Syn-STARTS datasets are a valid and trustworthy proxy for expert-authored benchmarks. This conclusion rests on two pillars of evidence: our synthetic cases were perceptually indistinguishable from expert-authored ones to clinical experts (\figureref{fig:3}), and LLM performance showed strong correlation across our synthetic Syn-STARTS datasets and expert-authored TRIAGE-adult dataset (Pearson’s $r = 0.92$, $p < 0.001$) (\figureref{fig:4}). This validation is critical because it confirms that our synthetic approach achieves parity with the gold standard in data quality and evaluation fidelity. Having shown this equivalence, the inherent advantages of our framework --- scalability that bypasses manual bottlenecks and a fully synthetic nature that eliminates patient privacy risks --- position it as a superior methodology. It matches the reliability of traditional methods while overcoming their most fundamental constraints, offering a more scalable, ethical, and powerful path for the rigorous evaluation of medical LLMs.

Second, our finding that dataset composition can directly influence model accuracy (\figureref{fig:5}) carries significant implications for evaluation design. We demonstrated that merely altering the distribution of triage tags caused statistically significant performance drops in certain models, proving that a single accuracy score on a static benchmark can be misleading. This vulnerability is explained by our model-specific tag sensitivity analysis (\figureref{fig:7}), which revealed that models have unique, tag-specific error patterns that are obscured by an aggregate score. A single metric, therefore, can conceal critical and heterogeneous failure modes. This underscores the need for evaluators to move beyond single-point metrics and instead justify the clinical and epidemiological representativeness of their benchmarks. Syn-STARTS provides the exact toolset required to perform these essential, multi-faceted sensitivity analyses.

Finally, our results demonstrate that scale is not merely a means to increase statistical power but a prerequisite for deeper diagnostic insights. We demonstrated that systematically increasing the dataset scale reduces evaluation variance (\figureref{fig:6}), thereby directly enhancing the stability and reproducibility of results --- a pervasive challenge in LLM research. This scale was also instrumental in uncovering fine-grained, model-specific error profiles, such as a model's tendency to confuse ``Immediate'' with ``Delayed'' cases (\figureref{fig:7}). Such insights, which are critical for safe deployment, are simply unattainable with smaller, resource-constrained test sets. Thus, scalability is not just about achieving a more precise score; it is what enables the transition from simplistic performance measurement to a comprehensive, diagnostic characterization of a model’s strengths and vulnerabilities.

In conclusion, the Syn-STARTS framework is more than a credible alternative to expert-authored datasets; it is a powerful and robust methodology. By enabling controllable, scalable, and fine-grained analysis, it provides a scientific instrument for deeply probing the behavior, vulnerabilities, and reliability of LLMs in clinically relevant scenarios. This work thus lays the groundwork for a more rigorous and trustworthy assessment of AI in medical high-stakes environments.

\subsection{Limitations and Future Directions}
While Syn-STARTS effectively assessed the capacity of LLMs to implement the START algorithm, our evaluation was limited to static and contextually independent cases.
Real-world MCIs, however, present a fundamentally different context where patients emerge from a single, evolving incident and are interconnected within that shared environment. Consequently, the effective application of LLMs in practical settings requires broader capabilities beyond mere tag assignment, including incident-specific judgment (e.g., assessing scene safety), triage command and control, and multi-modal reasoning to incorporate non-textual information. Assessing these complex, interconnected skills will necessitate the development of new frameworks capable of modeling the dynamic and context-rich environment characteristic of actual MCIs.

A further methodological consideration is the inherent tension between ensuring data quality and achieving narrative diversity (\appendixref{apd:chara} compares linguistic traits). As noted in \sectionref{sec:gen-and-val}, maintaining generated case validity requires a rigorous validation of narrative consistency, restricting expressions to align with vital signs. Expanding linguistic diversity to create more varied cases could, in turn, risk compromising the sufficiency of this validation process. Building upon the Syn-STARTS concept, a promising future direction is the development of a more granularly defined algorithm for case generation. Such a refinement could enable the construction of even higher-quality datasets, successfully balancing rich narrative diversity with unimpeachable data integrity.

Additionally, the application and expansion of the Syn-STARTS concept require the selection of the generator LLM and the precise design of prompts used within the framework. These factors are acknowledged as critical areas for future research.

A key future direction is to leverage our framework for an in-depth investigation into algorithmic fairness and bias. This encompasses the capability to examine traditional attribute biases by systematically controlling patient demographics, such as age and sex. Additionally, our framework facilitates an innovative exploration of narrative bias. This more subtle vulnerability involves testing whether an LLM’s clinical judgment is disproportionately influenced by a patient’s emotional state or communication style, independent of their primary clinical data. Investigating both demographic and these more nuanced narrative vulnerabilities is vital for the development of artificial intelligence systems that maintain objectivity and efficacy within the complex, high-pressure contexts of real-world medicine.

Furthermore, our concept of synthesizing benchmarks automatically represents a promising new direction for addressing a central challenge in medical AI: the scarcity of large, diverse, and well-annotated datasets for robust LLM evaluation. Our approach of generating synthetic, controllable case data presents a highly adaptable paradigm with the potential to resolve this critical data bottleneck. While demonstrated for triage, this method could be extended to evaluate a range of other established clinical decision-making algorithms, from diagnostic reasoning in internal medicine to treatment planning in oncology. Importantly, the process itself yields valuable insights into the data characteristics, such as scale and diversity, required for reliable LLM assessment. As such, this work offers not only a specific tool but also a scalable concept that addresses key data challenges and contributes to our understanding of how to rigorously evaluate LLMs in healthcare.

\newpage

\bibliography{references}

\appendix

\section{Prompt Templates}\label{apd:prompt}
This section presents the exact prompts used in our experiments (\figureref{fig:8,fig:9,fig:10}).
Figures \ref{fig:8} and \ref{fig:9} depict the prompts for generating Syn-STARTS cases, as described in \sectionref{sec:generation}. Specifically, \figureref{fig:8} provides an overview, while \figureref{fig:9} details the complete content.
\figureref{fig:10} shows the prompt for the triage tag prediction task (\sectionref{sec:task-eval}), which follows the format established in prior research \citep{kirch2024triage}.

\newsavebox{\genprompt}
\begin{lrbox}{\genprompt}
\begin{tcolorbox}[
          width=2.0\linewidth,
          colframe=royalblue,
          colback=royalblue!5!white,
          title=Prompt Template for Syn-STARTS Case Generation,
          fonttitle=\bfseries,
    ]
    You are a structured data generation bot. \\
        Your sole purpose is to output a single raw JSON object and nothing else.\\
        Do NOT act as a medical assistant. Do NOT add any conversational text or explanation.\\
        The JSON object you generate must represent a patient in a disaster scenario.
    
        The JSON format MUST strictly adhere to the following schema. Do NOT add, omit, or change any keys.
        \begin{itemize}
            \item \texttt{"triage\_tag"}: The requested tag (\textit{TAG\_COLOR}).
            \item \texttt{"patient\_description"}: A single, concise narrative string. \textbf{This description MUST include the patient's age, sex, the disaster situation, and a summary of the key vital signs that justify the triage tag.}
            \item \texttt{"vitals\_info"}: A nested JSON object. \textbf{This data must logically result in the requested triage\_tag and be reflected in the patient\_description.}
            \item The minimum information required for triage tag determination MUST be included.
            \item \textbf{You may ONLY use the following keys: }
            \begin{itemize}
                \item \texttt{"can\_walk"} (boolean)
                \item \texttt{"respirations"}: object with \texttt{"rate"}, \texttt{"initial\_breathing"}, and \texttt{"breathing\_after\_maneuver"}
                \item \texttt{"perfusion"}: object with \texttt{"radial\_pulse\_present"}, and \texttt{"capillary\_refill\_seconds"}
                \item \texttt{"mental\_status"}: object with \texttt{"obeys\_commands"}
                \item \textbf{Do not introduce any other keys into the \texttt{"vitals\_info"} object.}
            \end{itemize}
        \end{itemize}
        The values in the \texttt{"vitals\_info"} object MUST logically result in the requested \texttt{triage\_tag} according to the provided START algorithm.
        The minimum information required for triage tag determination MUST be included, but other information may or may not be present.
    
        Here is the algorithm for your reference: \textit{START\_DESCRIPTION}
        \\
        
        Below are some examples of correctly formatted scenarios that follow these rules: \textit{FEW\_SHOT\_EXAMPLES}
        \\
    
        Now, generate a patient scenario for the tag: \textit{TAG\_COLOR}
    \end{tcolorbox}
\end{lrbox}

\begin{figure*}[htbp]
    \floatconts
    {fig:8}
    {\caption{The prompt template used for Syn-STARTS case generation (\sectionref{sec:generation}). The \textit{TAG\_COLOR} placeholder is substituted with the desired ground truth tag for the case to be generated. For a detailed explanation of the START algorithm (\textit{START\_DESCRIPTION}) and the specific examples used for few-shot prompting (\textit{FEW\_SHOT\_EXAMPLES}), refer to \figureref{fig:9}.}}
    {
    \usebox{\genprompt}
    }
\end{figure*}

\newsavebox{\startalgo}
\begin{lrbox}{\startalgo}
\begin{tcolorbox}[
        width=2.0\linewidth,
        colframe=royalblue,
        colback=royalblue!5!white,
        title=Description of START Algorithm,
        fonttitle=\bfseries,
    ]
    The START (Simple Triage and Rapid Treatment) method is as follows:
        \begin{enumerate}
            \item \textbf{Can the patient walk?}
            \begin{itemize}
                \item Yes: Tag as GREEN.
                \item No: Proceed to Step 2.
            \end{itemize}
            \item \textbf{Check Respirations.}
            \begin{itemize}
                \item Not breathing: Open airway. If breathing starts, tag as RED. If still not breathing, tag as BLACK.
                \item Breathing: Proceed to Step 3.
            \end{itemize}
            \item \textbf{Check Respiratory Rate.}
            \begin{itemize}
                \item Over 30 breaths per minute: Tag as RED.
                \item Under 30 breaths per minute: Proceed to Step 4.
            \end{itemize}
            \item \textbf{Check Perfusion.}
            \begin{itemize}
                \item Capillary refill over 2 seconds OR no radial pulse: Tag as RED.
                \item Capillary refill under 2 seconds AND radial pulse present: Proceed to Step 5.
            \end{itemize}
            \item \textbf{Check Mental Status.}
            \begin{itemize}
                \item Cannot follow simple commands: Tag as RED.
                \item Can follow simple commands: Tag as YELLOW.
            \end{itemize}
        \end{enumerate}
    \end{tcolorbox}
\end{lrbox}

\newsavebox{\fewshot}
\begin{lrbox}{\fewshot}
\begin{tcolorbox}[
        width=2.0\linewidth,
        colframe=royalblue,
        colback=royalblue!5!white,
        title=Example Case Used for Few-shot Prompting,
        fonttitle=\bfseries,
    ]
        Now, generate a patient scenario for the tag: RED
        \begin{verbatim}
{
  "triage_tag": "Red",
  "patient_description": "44-year-old male with sharp trauma to neck.
  Capillary refill of four seconds and the patient is not following simple commands.
  The patient is dripping blood everywhere.
  You cannot see if it is pulsatile under the bandages.",
  "vitals_info": {
    "can_walk": false,
    "respirations": { "rate": 28 },
    "perfusion": { "capillary_refill_seconds": 4 },
    "mental_status": { "obeys_commands": false }
  }
}    
        \end{verbatim}

% \\
%     Now, generate a patient scenario for the tag: YELLOW\\
% \{
%   "triage\_tag": "Yellow",
%   "patient\_description": "50 y/o F RR 20 Capillary refill 2 seconds Obeys commands Sitting on shoulder of road complaining of dizziness",
%   "vitals\_info": \{
%     "can\_walk": false,
%     "respirations": \{ "rate": 24 \},
%     "perfusion": \{ "capillary\_refill\_seconds": 2 \},
%     "mental\_status": \{ "obeys\_commands": true \}
%   \}
% \}
% \\
%     Now, generate a patient scenario for the tag: GREEN\\
% \{
%   "triage\_tag": "Green",
%   "patient\_description": "13-year-old male with a bleeding finger. He is ambulatory. He states his finger hurts. No obvious deformity.",
%   "vitals\_info": \{
%     "can\_walk": true
%   \}
% \}
    \end{tcolorbox}
\end{lrbox}

\begin{figure*}[htbp]
\floatconts
    {fig:9}
    {\caption{The specific content of the START method explanation and the few-shot case examples used in the generation prompt.}}
    % \centering
    {
    \subfigure[Contents of \textit{START\_DESCRIPTION}]{\label{fig:9a}
      \usebox{\startalgo}}
    \\[1ex]
    \subfigure[Example of \textit{FEW\_SHOT\_EXAMPLES}]{\label{fig:9b}
      \usebox{\fewshot}}
    }
\end{figure*}

\newsavebox{\taskprompt}
\begin{lrbox}{\taskprompt}
    \begin{tcolorbox}[
        width=2.0\linewidth,
        colframe=royalblue,
        colback=royalblue!5!white,
        title=Prompt for Evaluation Tasks (\textit{FEW\_SHOT\_EXAMPLES}),
        fonttitle=\bfseries,
    ]
    You are a doctor working in the emergency room of a hospital.\\
    A colleague of yours runs into the emergency room and shouts: 'There has been a mass casualty incident at the nearby football stadium! A bomb has exploded. \\
    Hurry, we need to help!' You and your team rush to the scene, and divide roles.\\
    You must triage all patients. Determine whether a patient goes to MINOR, DELAYED, IMMEDIATE, or EXPECTANT/DECEASED zones.\\
    \\
    MINOR: victim with relatively minor injuries, status is unlikely to deteriorate over days.\\
    DELAYED: victim has serious and potentially life-threatening injuries, status is unlikely to deteriorate significantly over several hours.\\
    IMMEDIATE: victim requires medical attention within 60 minutes or less for survival, includes compromises to patient's airway, breathing, and circulation.\\
    EXPECTANT/DECEASED: victim is unlikely to survive given the severity of their injuries, level of available care, or both.\\
    \\
    To which category would you assign the following patient?\\
    Respond with a single JSON object containing ``reasoning'' (string) and ``action'' (string) fields. Make sure to only use `` '', NOT `'. Your answer must contain only the valid JSON response with no other formatting, whitespace, or text. Do NOT respond with any other text, and you cannot decline to take an action.\\
    Use the following format:\\  `\verb|{ "reasoning": "Because the patient... they should be assigned category ... ",|\\ \verb|"action": "MINOR" }|'
    \end{tcolorbox}
\end{lrbox}

\begin{figure*}[htbp]
    \floatconts
    {fig:10}
    {\caption{Prompt used during the triage tag prediction task execution (\sectionref{sec:task-eval}), quoted from \citet{kirch2024triage}. }}
    {
    \usebox{\taskprompt}
    }
\end{figure*}

\section{Summary of Corpus Construction}\label{apd:algo}
\algorithmref{alg:synstarts-gen} summarizes the entire corpus construction process, introduced in \sectionref{sec:corpus}, in pseudocode.

\begin{algorithm}
\floatconts
  {alg:synstarts-gen}%
  {\caption{Syn-STARTS Corpus Construction}}%
{
Given a set of target tags $T$ and a required number of samples per tag $N$:
\begin{enumerate*}
  \item Initialize an empty corpus, $C$.
  \item For each tag in $T$:
  \begin{enumerate*}
    \item Initialize a counter for validated samples to 0.
    \item While the counter is less than $N$, \textbf{apply the Syn-STARTS framework's generate-and-validate process}:
    \begin{enumerate*}
      \item[\textit{a.}] \label{step:generate} \textit{Generate} a candidate case (vitals, description) using Llama-3.1-70B-Instruct for the given tag.
      \item[\textit{b.}] Perform a three-step \textit{validation} on the candidate:
      \begin{enumerate*}
        \item[\textit{i.}] \textbf{START Consistency}
        \item[\textit{ii.}] \textbf{Medical Plausibility}
        \item[\textit{iii.}] \textbf{Textual Consistency}
      \end{enumerate*}
      \item[\textit{c.}] If the candidate passes all three validation steps:
      \begin{enumerate*}
        \item[] Add the case to $C$ and increment the counter.
      \end{enumerate*}
      otherwise
      \begin{enumerate*}
        \item[] Discard the candidate and return to step~\ref{step:generate}.
      \end{enumerate*}
    \end{enumerate*}
  \end{enumerate*}
  \item Return the final validated corpus $C$, where $|C|=4N$.
\end{enumerate*}}%
\end{algorithm}

\section{Example of Syn-STARTS Case} \label{apd:example-cases}
\figureref{fig:examples} provides concrete examples of Syn-STARTS cases generated by our framework (\sectionref{sec:case}).

\newsavebox{\cardg}
\begin{lrbox}{\cardg}
    \definecolor{SGreen}{RGB}{40, 160, 40}
    \begin{tcolorbox}[
        % enhanced,
        colframe=SGreen,
        colback=SGreen!5,
        fonttitle=\color{white}\sffamily\bfseries,
        colbacktitle=SGreen,
        title={Example: Green 1}
        ]
        22-year-old female with\\ minor scratches on her face and arms. \\
        She is walking around and talking to others, complaining only of minor pain.
    \end{tcolorbox}
\end{lrbox}

\newsavebox{\cardgg}
\begin{lrbox}{\cardgg}
    \definecolor{SGreen}{RGB}{40, 160, 40}
    \begin{tcolorbox}[
        % enhanced,
        colframe=SGreen,
        colback=SGreen!5,
        fonttitle=\color{white}\sffamily\bfseries,
        colbacktitle=SGreen,
        title={Example: Green 2}
        ]
        25-year-old female, conscious and ambulatory, with minor scrapes on her knees after falling while trying to evacuate the building during the fire. She is walking around and talking with no signs of distress.
    \end{tcolorbox}
\end{lrbox}

\newsavebox{\cardy}
\begin{lrbox}{\cardy}
    \definecolor{SYellow}{RGB}{240, 190, 30}
    \begin{tcolorbox}[
        % enhanced,
        colframe=SYellow,
        colback=SYellow!5,
        fonttitle=\color{black}\sffamily\bfseries,
        colbacktitle=SYellow,
        title={Example: Yellow 1}
        ]
        70-year-old male with a crushed leg. He is unable to walk, but can obey commands.\\Respiratory rate is 22 and capillary refill is 1 second.
    \end{tcolorbox}
\end{lrbox}

\newsavebox{\cardyy}
\begin{lrbox}{\cardyy}
    \definecolor{SYellow}{RGB}{240, 190, 30}
    \begin{tcolorbox}[
        % enhanced,
        colframe=SYellow,
        colback=SYellow!5,
        fonttitle=\color{black}\sffamily\bfseries,
        colbacktitle=SYellow,
        title={Example: Yellow 2}
        ]
        75-year-old female who was trapped under debris, now freed. RR 18. She cannot stand or walk due to left leg injury and severe pain but is alert and obeys commands.
    \end{tcolorbox}
\end{lrbox}

\newsavebox{\cardr}
\begin{lrbox}{\cardr}
    \definecolor{SRed}{RGB}{220, 50, 50}
    \begin{tcolorbox}[
        % enhanced,
        colframe=SRed,
        colback=SRed!5,
        fonttitle=\color{white}\sffamily\bfseries,
        colbacktitle=SRed,
        title={Example: Red 1}
        ]
        29-year-old female who was pinned under debris. She is not breathing. After opening her airway, she begins breathing but her respiratory rate is 32 breaths per minute.
    \end{tcolorbox}
\end{lrbox}

\newsavebox{\cardrr}
\begin{lrbox}{\cardrr}
    \definecolor{SRed}{RGB}{220, 50, 50}
    \begin{tcolorbox}[
        % enhanced,
        colframe=SRed,
        colback=SRed!5,
        fonttitle=\color{white}\sffamily\bfseries,
        colbacktitle=SRed,
        title={Example: Red 2}
        ]
        25 y/o M with severe head trauma, bleeding from the ears and nose.\\ Respiratory rate is 38 and the patient is confused and disoriented.
    \end{tcolorbox}
\end{lrbox}

\newsavebox{\cardb}
\begin{lrbox}{\cardb}
    \definecolor{SBlack}{RGB}{50, 50, 50}
    \begin{tcolorbox}[
        % enhanced,
        colframe=SBlack,
        colback=SBlack!5,
        fonttitle=\color{white}\sffamily\bfseries,
        colbacktitle=SBlack,
        title={Example: Black 1}
        ]
        75-year-old female, found unresponsive with severe head trauma and not breathing.\\CPR was initiated, but she did not start breathing after opening the airway.
    \end{tcolorbox}
\end{lrbox}

\newsavebox{\cardbb}
\begin{lrbox}{\cardbb}
    \definecolor{SBlack}{RGB}{50, 50, 50}
    \begin{tcolorbox}[
        % enhanced,
        colframe=SBlack,
        colback=SBlack!5,
        fonttitle=\color{white}\sffamily\bfseries,
        colbacktitle=SBlack,
        title={Example: Black 2}
        ]
        67-year-old female unresponsive, not breathing, after being crushed under debris.\\Attempted to open airway but no breathing resumed.
    \end{tcolorbox}
\end{lrbox}

\begin{figure*}[hbtp]
\thisfloatsetup{valign=c}
\floatconts
  {fig:examples}
  {\caption{Examples of Syn-STARTS cases. Two examples are provided for each of the four START triage categories.}}
  {
  \subfigure[Case with Green Tag (Minor) \#1]{\label{fig:g1}\usebox{\cardg}}
  \subfigure[Case with Green Tag (Minor) \#2]{\label{fig:g2}\usebox{\cardgg}}
  \\[4ex]
  \subfigure[Case with Yellow Tag (Delayed) \#1]{\label{fig:y1}\usebox{\cardy}}
  \subfigure[Case with Yellow Tag (Delayed) \#2]{\label{fig:y2}\usebox{\cardyy}}
  \\[4ex]
  \subfigure[Case with Red Tag (Immediate) \#1]{\label{fig:r1}\usebox{\cardr}}
  \subfigure[Case with Red Tag (Immediate) \#2]{\label{fig:r2}\usebox{\cardrr}}
  \\[4ex]
  \subfigure[Case with Black Tag (Deceased) \#1]{\label{fig:b1}\usebox{\cardb}}
  \subfigure[Case with Black Tag (Deceased) \#2]{\label{fig:b2}\usebox{\cardbb}}
  }
\end{figure*}

\section{Datasets' Characteristics}\label{apd:chara}
We conducted a comparative analysis of linguistic features between the TRIAGE-adult dataset and our Syn-STARTS datasets to validate the structural properties of the synthetic data. Our analysis, summarized in \tableref{tab:chara}, focused on key metrics such as narrative length and vocabulary size. Furthermore, \figureref{fig:chara} illustrates the word length distributions of the datasets.

\begin{figure*}[htbp]
\floatconts
  {fig:chara}
  {\caption{Comparison of narrative length distributions between (a) TRIAGE-adult dataset and (b) Syn-STARTS datasets with matched tag distribution ($n=54$; $\{18, 11, 22, 3\}$). (b) shows the averaged scores of ten datasets and error bars indicating the standard deviation.}}
  {
    \subfigure[TRIAGE-adult]{\label{fig:12a}
      \includegraphics[width=0.8\linewidth]{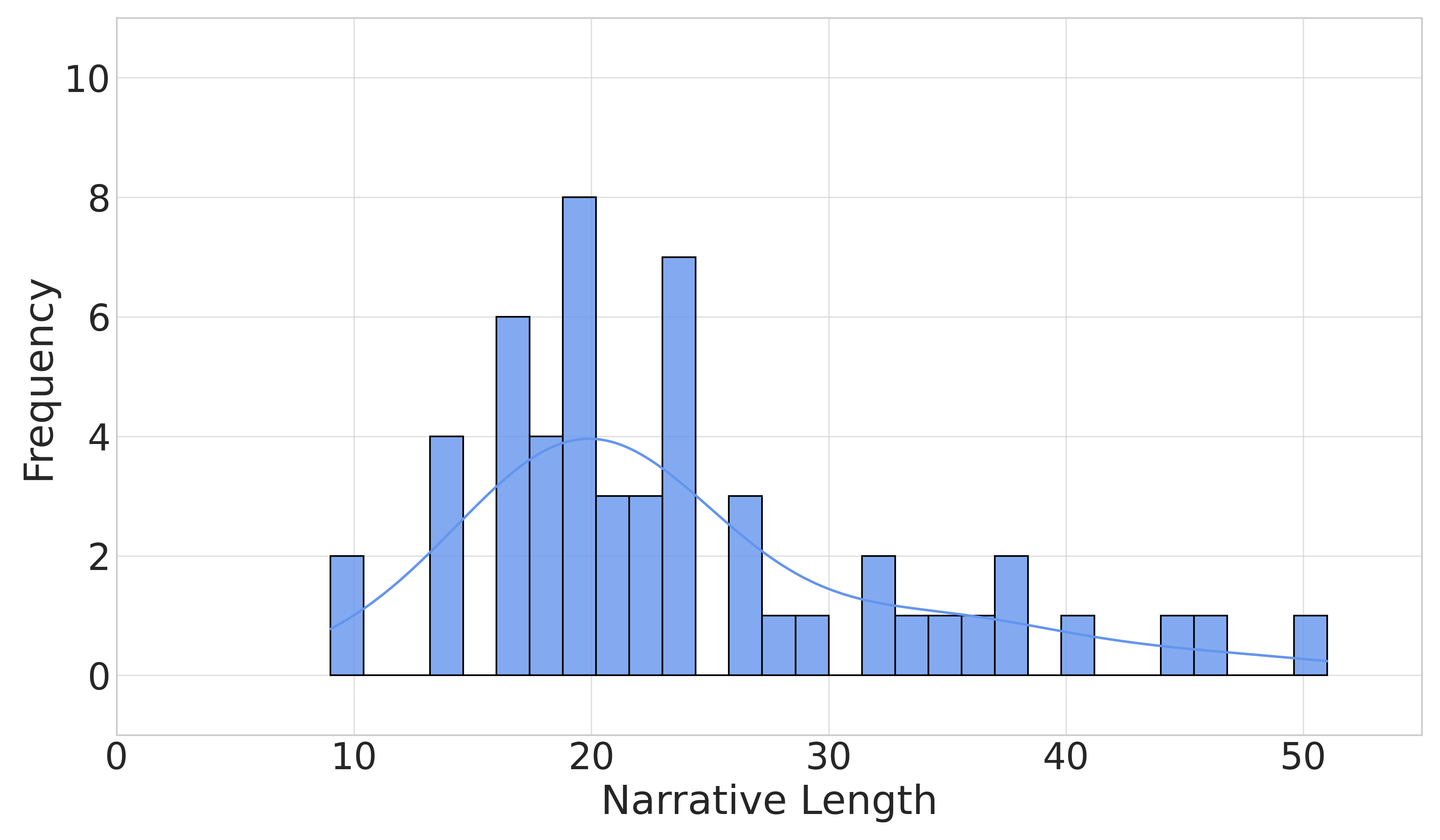}}
    \\
    \subfigure[Syn-STARTS datatsets]{\label{fig:12b}
      \includegraphics[width=0.8\linewidth]{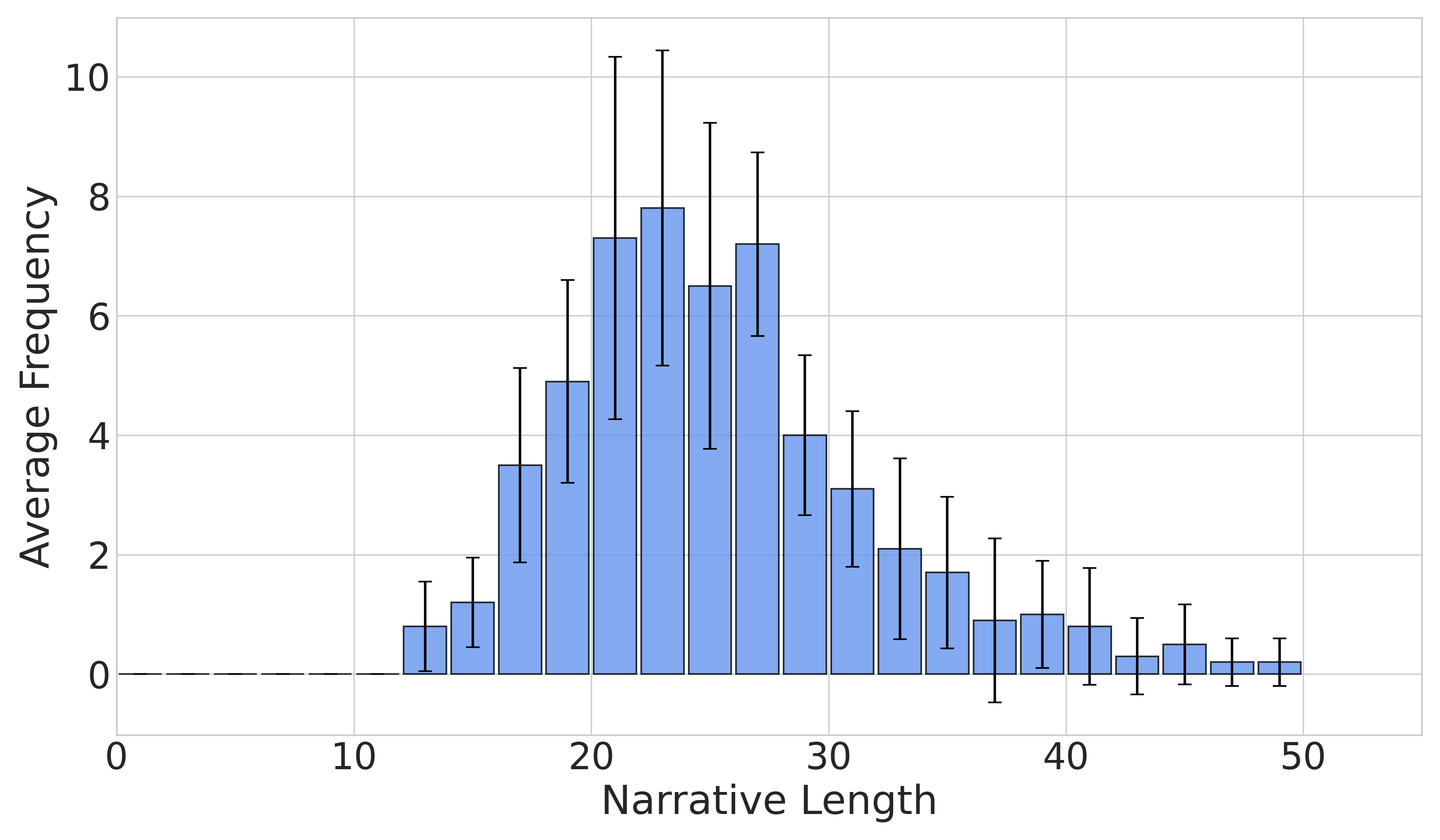}}
  }
\end{figure*}

\begin{table*}[htbp]
\floatconts
 {tab:chara}
 {\caption{Linguistic features for the TRIAGE-adult dataset and Syn-STARTS datasets. Values for Syn-STARTS are presented as mean ± standard deviation across the generated datasets.}}
 {\begin{tabular}{ccc}
    \toprule
    & TRIAGE-adult Dataset & Syn-STARTS Datasets\\
    \midrule
    Average of Narrative Length & 23.67 & 24.90 $\pm$ 0.89\\
    Vocabulary Size & 310 & 186.90 $\pm$ 14.14\\
    \bottomrule
 \end{tabular}}
\end{table*}

\section{Supplementary Results}\label{apd:tables}
\tableref{tab:results} presents the detailed numerical accuracy scores for our experiments. These results form the basis for the figures shown in \sectionref{sec:results}.

\begin{itemize}
    \item \tableref{tab:1a} lists the accuracy scores for the \textbf{Assessment of Fidelity}, as discussed in Section \ref{sec:fidelity} and visualized in Figure \ref{fig:4}.
    \item \tableref{tab:1b} shows the results for the assessment of the \textbf{Effect of Tag Distribution}, corresponding to the analysis in Section \ref{sec:tagdist} and Figure \ref{fig:5}.
    \item \tableref{tab:1c} details the outcomes for the assessment of the \textbf{Effect of Dataset Scale}, as described in Section \ref{sec:scale} and shown in Figure \ref{fig:6}.
\end{itemize}

\begin{table*}[htbp]
\floatconts
 {tab:results}
 {\caption{Detailed results scores. (a): Results of \textbf{Assessment of Fidelity} (\sectionref{sec:fidelity}, visualized in \figureref{fig:4}); Model accuracy on (A) the expert-authored TRIAGE-adult dataset versus (B) Syn-STARTS datasets with matched tag distributions ($n=54$; $\{18, 11, 22, 3\}$). (b): Results of assessment of \textbf{Effect of Tag Distribution} (\sectionref{sec:tagdist}, visualized in \figureref{fig:5}); Accuracy of six models across Syn-STARTS datasets with (A) ``TRIAGE adult'' ($n=54$; $\{18, 11, 22, 3\}$) versus (B) uniform ($n=56$; $\{14, 14, 14, 14\}$) tag distributions. (c): Results of assessment of \textbf{Effect of Dataset Scale} (\sectionref{sec:scale}, visualized in \figureref{fig:6}): Model accuracy and variance across Syn-STARTS datasets of increasing scale.}}
 {%
   \subtable[\textbf{Assessment of Fidelity}]{%
     \label{tab:1a}%
     \begin{tabular}{ccc}
  \toprule
  \bfseries Models & \multicolumn{2}{c}{\bfseries Datasets} \\
  \cmidrule(lr){2-3}
  & \bfseries (A) & \bfseries (B) \\
  \midrule
  Mistral-7B & 0.29 & 0.21 $\pm$ 0.03\\
  Mixtral-8x22B & 0.64 & 0.86 $\pm$ 0.01\\
  Claude3-Haiku & 0.57 & 0.58 $\pm$ 0.05\\
  Claude-Opus-4 & 0.66 & 0.92 $\pm$ 0.02\\
  GPT-3.5 & 0.57 & 0.85 $\pm$ 0.03\\
  GPT-4 & 0.72 & 0.85 $\pm$ 0.02\\
  \bottomrule
  \end{tabular}
   }\qquad
   \subtable[Assessment of \textbf{Effect of Tag Distribution}]{%
     \label{tab:1b}%
     \begin{tabular}{cccc}
  \toprule
  \bfseries Models & \multicolumn{2}{c}{\bfseries Datasets} & \bfseries p-value\\
  \cmidrule(lr){2-3}
  & \bfseries (A) & \bfseries (B) \\
  \midrule
  Mistral-7B & 0.21 $\pm$ 0.03 & 0.20 $\pm$ 0.05 & 0.67\\
  Mixtral-8x22B & 0.86 $\pm$ 0.01 & 0.86 $\pm$ 0.03 & 0.53\\
  Claude3-Haiku & 0.58 $\pm$ 0.05 & 0.53 $\pm$ 0.04 & 0.08\\
  Claude-Opus-4 & 0.92 $\pm$ 0.02 & 0.92 $\pm$ 0.02 & 0.90\\
  GPT-3.5 & 0.85 $\pm$ 0.03 & 0.78 $\pm$ 0.04 & \textbf{0.00}\\
  GPT-4 & 0.85 $\pm$ 0.02 & 0.81 $\pm$ 0.03 & \textbf{0.04}\\
  \bottomrule
  \end{tabular}
   }\\[2em]
   \subtable[Assessment of \textbf{Effect of Dataset Scale}]{%
     \label{tab:1c}%
     \begin{tabular}{cccccc}
  \toprule
  \bfseries Models & \multicolumn{2}{c}{\bfseries Datasets}\\
  \cmidrule(lr){2-6}
  & \bfseries $n=12$ & \bfseries $n=20$ & \bfseries $n=56$ & \bfseries $n=100$ & \bfseries $n=200$ \\
  \midrule
  Mistral-7B & 0.18 $\pm$ 0.0765 & 0.18 $\pm$ 0.0474 & 0.20 $\pm$ 0.0596 & 0.21 $\pm$ 0.0258 & 0.22 $\pm$ 0.0193\\
  Mixtral-8x22B & 0.83 $\pm$ 0.1039 & 0.83 $\pm$ 0.0474 & 0.86 $\pm$ 0.0383 & 0.84 $\pm$ 0.0362 & 0.83 $\pm$ 0.0240\\
  Claude3-Haiku & 0.51 $\pm$ 0.1024 & 0.49 $\pm$ 0.0724 & 0.53 $\pm$ 0.0474 & 0.53 $\pm$ 0.0378 & 0.52 $\pm$ 0.0345\\
  Claude-Opus-4 & 0.93 $\pm$ 0.0527 & 0.93 $\pm$ 0.0474 & 0.92 $\pm$ 0.0312 & 0.92 $\pm$ 0.0218 & 0.92 $\pm$ 0.0172\\
  GPT-3.5 & 0.76 $\pm$ 0.1290 & 0.76 $\pm$ 0.0567 & 0.78 $\pm$ 0.0510 & 0.79 $\pm$ 0.0270 & 0.79 $\pm$ 0.0227\\
  GPT-4 & 0.78 $\pm$ 0.0582 & 0.80 $\pm$ 0.0283 & 0.81 $\pm$ 0.0405 & 0.82 $\pm$ 0.0200 & 0.82 $\pm$ 0.0181\\
  \bottomrule
  \end{tabular}
   }
 }
\end{table*}

\section{Further Analysis by Tags}\label{apd:tags}
\figureref{fig:11} illustrates the standard deviation of accuracy for each tag across different dataset scales ($n = 12, 20, 56, 100, \text{and } 200$). As shown, the standard deviation for all four tags exhibits a clear downward trend as the scale of the datasets increases. This trend mirrors the overall results presented in \figureref{fig:6}.

\begin{figure*}[htbp]
\floatconts
  {fig:11}
  {\caption{Standard deviation of model accuracy across dataset scales, shown for each tag. The consistent downward trend shows that increasing dataset scale reduces variance in accuracy estimates, for each tag as well.}}
  {
    \subfigure[For Green Tag]{\label{fig:6b}
      \includegraphics[width=0.48\linewidth]{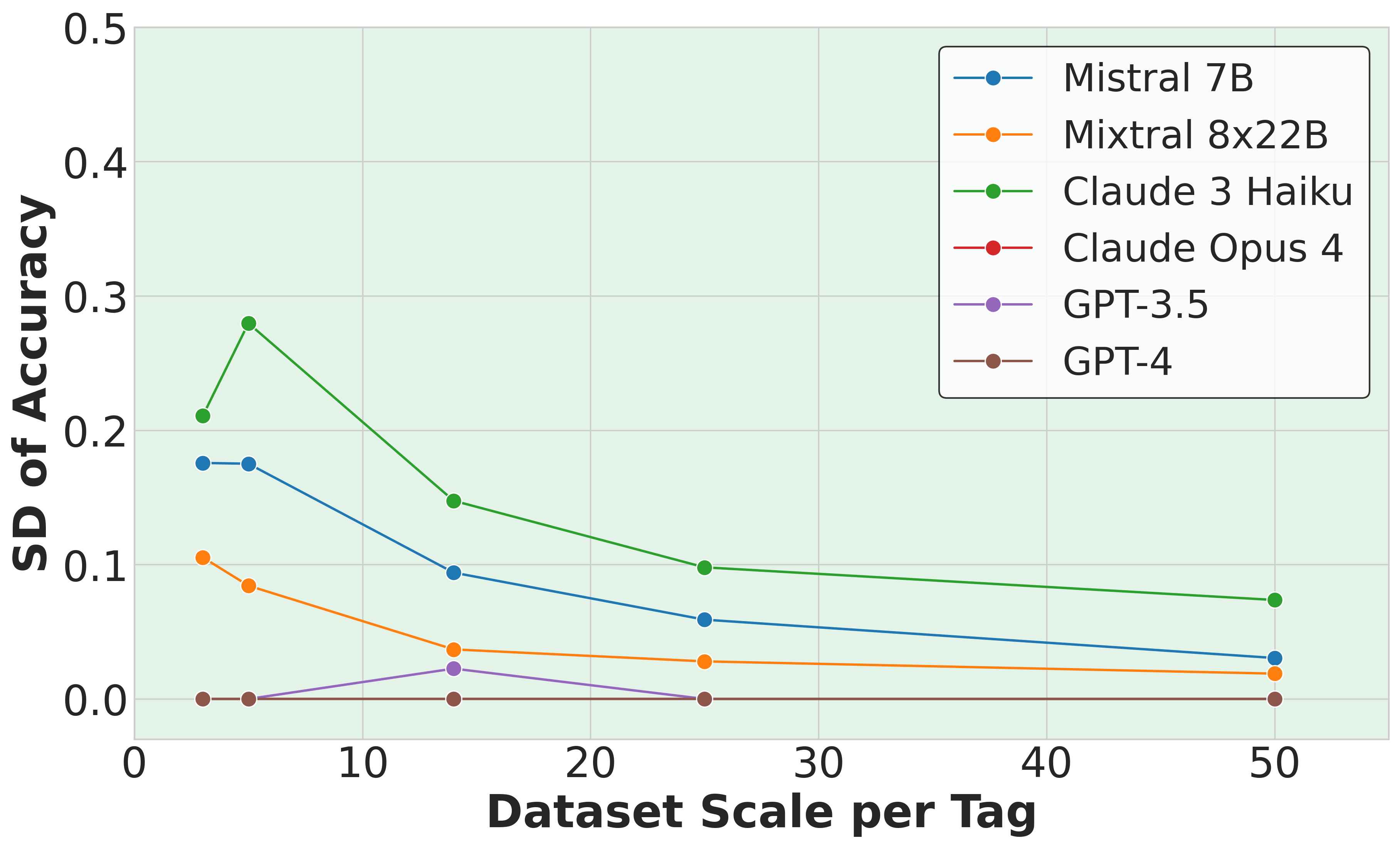}}
    \subfigure[For Yellow Tag]{\label{fig:6c}
      \includegraphics[width=0.48\linewidth]{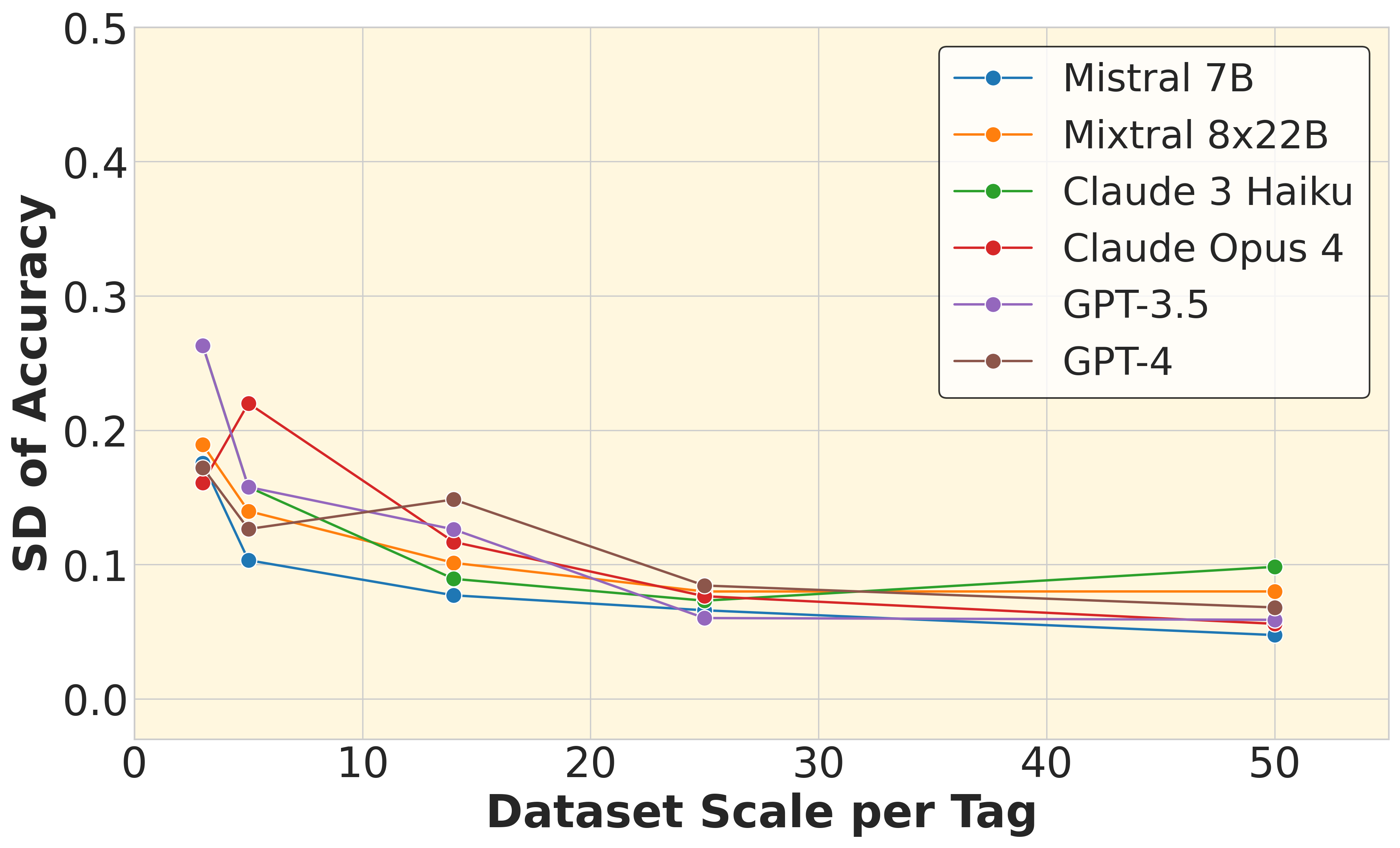}}
    \\[2em]
    \subfigure[For Red Tag]{\label{fig:6d}
      \includegraphics[width=0.48\linewidth]{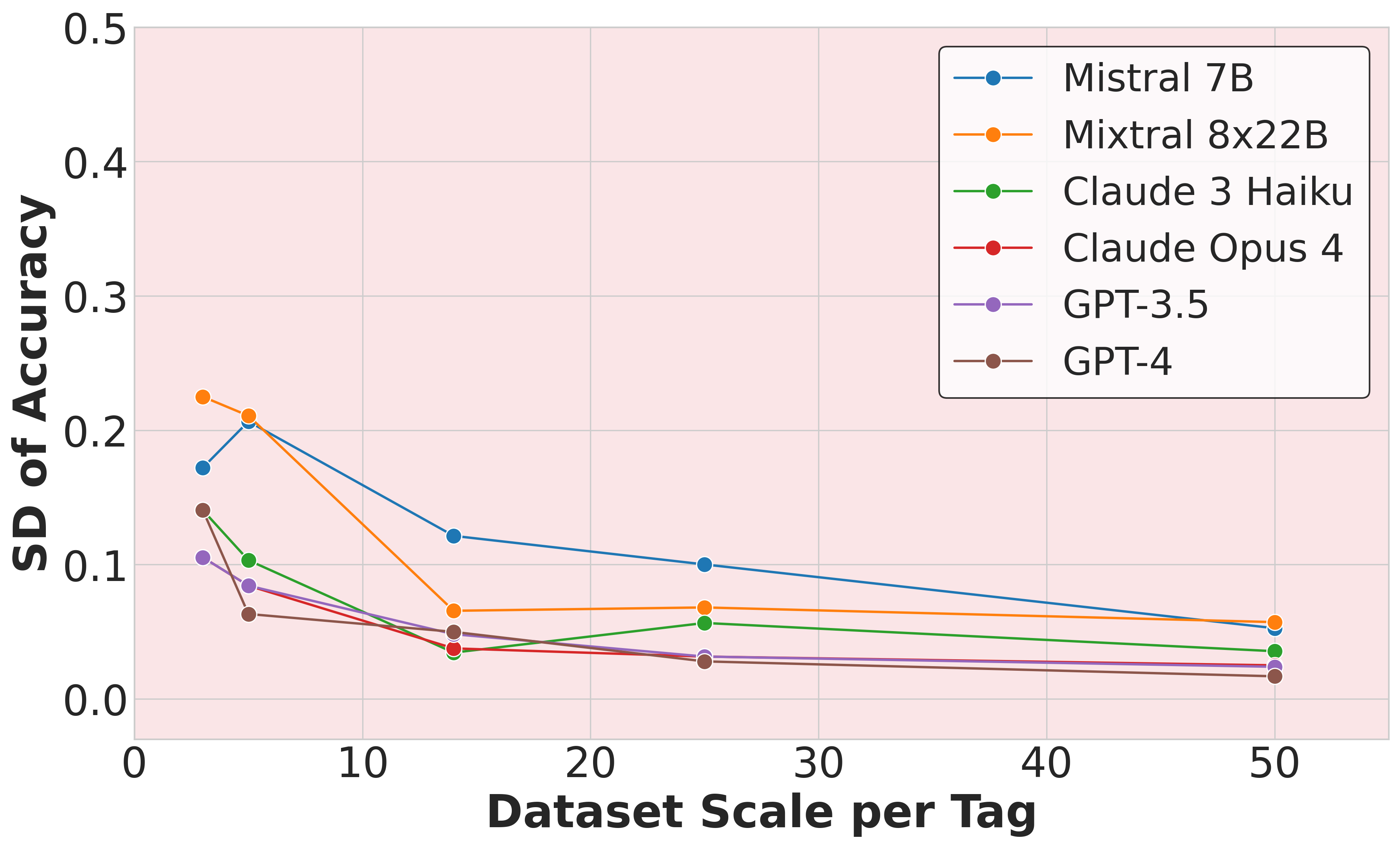}}
    \subfigure[For Black Tag]{\label{fig:6e}
      \includegraphics[width=0.48\linewidth]{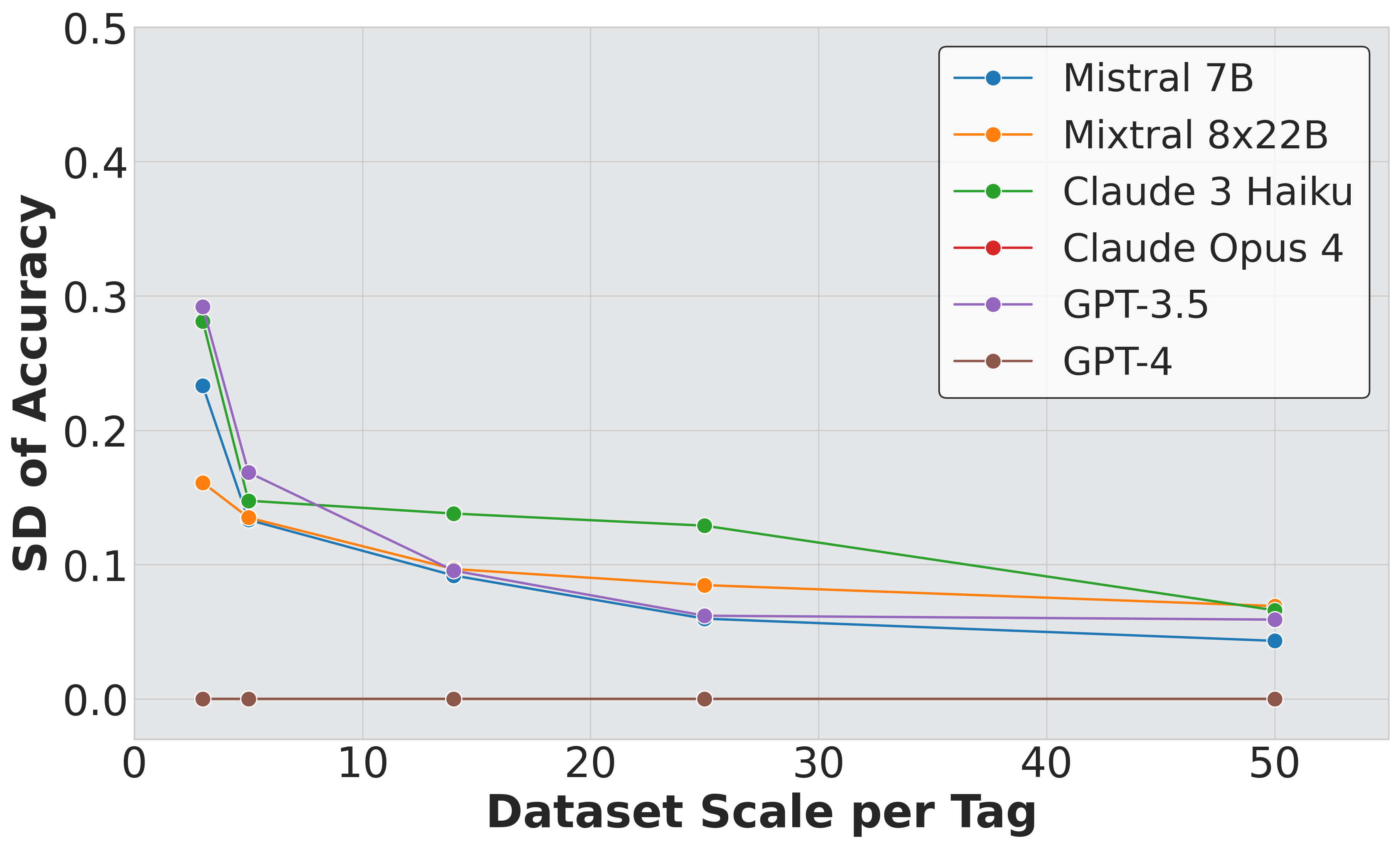}}
  }
\end{figure*}

\end{document}